\pdfoutput=1
\documentclass[10pt]{article}
\usepackage[T1]{fontenc}
\usepackage[utf8]{inputenc}
\usepackage{lmodern}
\usepackage[letterpaper,margin=0.8in]{geometry}
\usepackage{microtype}
\usepackage{amsmath,amssymb}
\usepackage{array,booktabs,longtable,tabularx}
\usepackage{enumitem}
\usepackage{multicol}
\usepackage[table]{xcolor}
\usepackage{float}
\usepackage{caption}
\usepackage{tikz}
\usetikzlibrary{arrows.meta,positioning,calc,decorations.pathreplacing}
\usepackage[colorlinks=true,linkcolor=linkc,citecolor=linkc,urlcolor=linkc]{hyperref}
\hypersetup{pdftitle={Anatomy of an Agent Harness: Decisions, Invariants and Corner Cases},pdfauthor={Resham Joshi}}

\definecolor{linkc}{HTML}{1F4E79}
\definecolor{tagT}{HTML}{2E6B4F}
\definecolor{tagB}{HTML}{9A6A12}
\definecolor{tagO}{HTML}{5A5A5A}
\definecolor{tbfill}{HTML}{E8EEF4}
\definecolor{bndc}{HTML}{A23B2A}
\definecolor{cardrule}{HTML}{9AA7B4}
\definecolor{appl}{HTML}{55606B}

\makeatletter
\newcommand{\needspace}[1]{\par\penalty-100\begingroup\setlength{\dimen@}{#1}%
  \dimen@ii\pagegoal\advance\dimen@ii-\pagetotal
  \ifdim\dimen@>\dimen@ii\ifdim\dimen@ii>\z@\vfil\fi\penalty-\@M\fi\endgroup}
\makeatother

\newcommand{\TagT}{{\sffamily\bfseries\small TESTED}}
\newcommand{\TagB}{{\sffamily\bfseries\small BELIEVED}}
\newcommand{\TagO}{{\sffamily\bfseries\small OPEN}}
\newcommand{\Tst}{{\sffamily\bfseries\small Tested}}
\newcommand{\Bel}{{\sffamily\bfseries\small Believed}}
\newcommand{\Opn}{{\sffamily\bfseries\small Open}}

\newcommand{\gl}[1]{\emph{#1}\textsuperscript{\hyperref[app:glossary]{\scriptsize\sffamily G}}}

\newcommand{\card}[4]{\par\addvspace{5pt}\needspace{5\baselineskip}%
  \noindent\textcolor{cardrule}{\rule{\linewidth}{0.4pt}}\par\nobreak\vspace{1pt}%
  \noindent\phantomsection\label{card:#1}{\bfseries Card #1\quad #2}\hfill{\footnotesize\sffamily\textcolor{appl}{#4}}\quad #3\par\nobreak\vspace{1pt}}
\newcommand{\plain}[1]{\noindent{\small\emph{In plain words:} #1}\par\nobreak}
\newenvironment{fields}{\small\begin{description}[leftmargin=1.2em,itemsep=0.6pt,topsep=1pt,parsep=0pt,font=\normalfont\sffamily\bfseries]}{\end{description}}
\newenvironment{opts}{\begin{description}[leftmargin=1em,itemsep=0.4pt,topsep=0pt,partopsep=0pt,parsep=0pt,font=\normalfont\itshape]}{\end{description}}
\newcommand{\inv}[1]{\textit{#1}}
\newcommand{\cd}[1]{\hyperref[card:#1]{\textsf{#1}}}

\DeclareRobustCommand{\bnd}[1]{\tikz[baseline=(b.base)]\node[circle,draw=bndc,text=bndc,inner sep=0.6pt,font=\footnotesize\bfseries\sffamily](b){#1};}

\title{Anatomy of an Agent Harness:\\Decisions, Invariants and Corner Cases}
\author{Resham Joshi\\\small\texttt{resham@codeburn.app}}
\date{}

\begin{document}
\maketitle

\begin{abstract}
\noindent An agent harness is the program around a language model that turns its replies into actions: it builds each request, runs the tools the model asks for, decides what the model may do without asking, stores the conversation, recovers from failures, and shows a person what happened. Most of these decisions are made in code and rarely written down, so each harness meets the same corner cases again. This paper writes them down as invariants to test. It gives a system model with trust boundaries; 30 decision cards in six groups, each with options and costs, an invariant, the corner cases that break it, a test and an evidence tag; six cause classes for failures; a test method built on scripted providers, recorded behaviour and planted changes; and a checklist. The cards come from building one desktop harness, Crewroom, the author's own, and from manually testing ten other harnesses and studying open-source harnesses as reference; some patterns in the cards were adopted from those, so the catalogue is not independent of them. Of the 83 invariant clauses, 56 are tagged Tested, 12 Believed and 15 Open. The checklist is applied to the author's harness as one worked example, by its builder with no second rater: of its 30 items, 11 hold, 14 do not and 5 were not tested (Section~\ref{sec:cross}). It is not a ranking. The paper makes no claims about answer quality or cost or other platforms, and does not evaluate or rank other harnesses.
\end{abstract}

\section{Scope and terms}\label{sec:scope}

A language model on its own only writes text. To let it read files, run commands or open web pages, someone has to write a program that sends the model a request, reads the reply, carries out the actions the reply asks for, sends the results back, and repeats until the work is done. That program is the \emph{harness}. The harness decides what happens when a reply is cut off, the network drops, the model asks to delete a file, or a person presses Stop. This paper lists those decisions, so that a builder can check a harness against them and a reader who does not program can see what is at stake.

\paragraph{Contribution.} The paper writes each harness decision as an invariant to test. Each \emph{decision card} records one decision, its options with their gains and costs, an \emph{invariant}, the corner cases that break it, a test, and how strong the evidence is. None of these decisions is new on its own. What is new is writing each one down with an invariant and a test, across the \emph{turn}, failures, the \emph{prompt}, state, permission and display, and tagging every claim by the evidence behind it. The checklist is then applied to our own harness as one worked example (Section~\ref{sec:cross}). Not claimed: that any option is best for every harness, that the 30 cards are complete, that any harness meets the invariants, or anything about answer quality or cost.

\paragraph{What a harness is and is not.} Here a harness is everything between a person (or a schedule) and a model \gl{provider} that is needed to run a multi-step agent: the \emph{engine} that runs the turn loop, the provider layer that speaks each \emph{wire format}, the tools, the permission layer and \emph{sandbox}, the stored \emph{transcript}, the \emph{host} process that starts engines and connects them to a window, and the display. It is not the model, the provider's service, the internals of individual tools, or a benchmark. Prompt text is in scope only where the harness assembles it from parts (Section~\ref{sec:told}).

\paragraph{Terms.} A \emph{turn} is everything between one message that starts work and the harness giving control back. A \emph{step} is one request to the model plus the tool runs that follow it. A \emph{tool call} is the model's request to run a tool, identified by a call id; its \emph{result} is what the harness sends back. An \emph{end reason} is the recorded cause of a turn ending. A turn is \emph{watched} when a person is there to answer questions, and unwatched otherwise. The \emph{checker} decides before a command runs whether it may run without asking; an \emph{approval request} asks the person. A \emph{prompt} is always text sent to the model, never a question to the person. \emph{Compaction} replaces old conversation with a \emph{summary} so it fits the model's \emph{context limit}. A \emph{helper} is a second agent started by the first. A \emph{model family} is a group of models trained alike and served alike. Terms marked {\scriptsize\sffamily G} at first use are defined in Appendix~\ref{app:glossary}.

\paragraph{What the paper rests on.} The cards come from building one desktop harness, Crewroom, the author's own; the rest of the paper calls it \emph{our harness}. It runs models from several families through several wire formats. It runs models from several families through several wire formats. The other source is manual testing and observation of ten other harnesses, with open-source harnesses also studied as reference (Section~\ref{sec:compare}). Some patterns and rules described in the cards were adopted from those harnesses, so the catalogue is not independent of them. Appendix~\ref{app:evidence} gives the evidence per card without names. Of the 30 cards, 24 rest on \emph{scripted providers} alone in the recorded evidence (T1 and T4 also cite live sightings from the author's notes); of the other six, five (F1, F2, F3, F4 and P8) also reach the screen of the installed application, the real sandbox or a real model, and P3 was checked in the real sandbox only (Table~\ref{tab:cards}). The cards come from failures met in our harness, so the list leans toward its design, and ``Tested'' means ``tested in our harness'', not ``true of all harnesses''. Section~\ref{sec:open} states the limits of this evidence.

\paragraph{How to read a card.} Each card (T1, F3, P5, \ldots) gives a sentence in plain words; the \textbf{Decision} and why it is forced; \textbf{Options}, each choice with its gain and cost; the \textbf{Invariant}, in lettered clauses a test can check; \textbf{Corner cases} that break it; a \textbf{Test}; and for permission cards a \textbf{Cost and limit}, what the rule does not prevent. The Invariant is meant as the universal part: it should hold in every harness the heading names, whichever option is taken. That scope is reasoned (\Bel); invariants that are our own design choices are worded as choices. Options are implementation choices, and a number inside them is a value chosen in our harness unless the text gives it to another harness tested.

Each clause carries a strength tag. \Tst: a test or live check of the clause existed in our harness and passed when the evidence was recorded; for a clause about every case, this means no counterexample in the cases tried. \Bel: reasoned, seen once, observed in other harnesses, tested only in parts never confirmed as a whole, or tested with no passing run recorded. \Opn: unresolved, or found false in our harness and not yet fixed. Tags were taken from the author's notes on the test records, not from fresh runs, except that a clause the runs of Section~\ref{sec:cross} found false in our harness is Open. A run can only lower a tag; a clause becomes Tested only through a test in the suite. The tag in a card's heading is that of clause (a), its core. By that tag 22 cards are Tested, 3 Believed (S4, P3, P5) and 5 Open (C1, P1, P2, P4, P9). Of all 83 clauses, 56 are Tested, 12 Believed and 15 Open: T3(b), T5(b), F1(b), F2(b), F2(d), F5(b), C1(a), S1(d), S2(d), S2(e), P1(a), P2, P4, P9(a) and V1(d). An option or corner case weaker than its card carries its own tag. The tags describe our harness only; the verdicts of the self-assessment stand beside them in Table~\ref{tab:cards}.

\section{System model}\label{sec:model}

\begin{figure}[tb]
\centering
\begin{tikzpicture}[font=\small,
  box/.style={draw,rounded corners=2pt,align=center,minimum height=10mm,inner sep=3pt,text width=2.7cm},
  sbox/.style={draw,rounded corners=2pt,align=center,minimum height=10mm,inner sep=2pt,text width=1.25cm,font=\footnotesize},
  tb/.style={fill=tbfill},
  arr/.style={-{Stealth[length=2mm]},semithick},
  darr/.style={{Stealth[length=2mm]}-{Stealth[length=2mm]},semithick},
  bn/.style={circle,draw=bndc,fill=white,text=bndc,inner sep=0.7pt,font=\footnotesize\bfseries\sffamily}]
\node[box,tb] (person) at (0,0) {Person and window};
\node[box] (ext) at (4.4,0) {Extensions\\\footnotesize skills, hooks, imported settings};
\node[box] (proj) at (8.8,0) {Project files\\\footnotesize rules, settings, code};
\node[box,tb] (host) at (0,-2.3) {Host process\\\footnotesize starts engines, routes messages};
\node[box,tb,minimum height=13mm] (engine) at (4.4,-2.3) {\textbf{Engine}\\turn loop};
\node[box,tb] (prov) at (8.8,-2.3) {Provider layer\\\footnotesize one adapter per wire format};
\node[box] (models) at (13.0,-2.3) {Model providers};
\node[sbox,font=\footnotesize] (others) at (-0.72,-4.7) {Other agents};
\node[sbox,font=\footnotesize] (sched) at (0.72,-4.7) {Schedules, webhooks};
\node[box,tb] (log) at (4.4,-4.7) {Stored transcript\\\footnotesize append-only log};
\node[box,tb,text width=2.9cm] (perm) at (8.8,-4.7) {Permission layer\\and sandbox};
\node[box] (tools) at (13.0,-4.7) {Tools\\\footnotesize shell, files, web, tool servers};
\draw[darr] (person) -- node[right,font=\footnotesize,align=left]{messages (person's mark),\\approval requests, answers} (host);
\draw[darr] (host) -- node[above,font=\footnotesize]{requests} node[below,font=\footnotesize]{events} (engine);
\draw[arr] (engine) -- node[left,font=\footnotesize]{append} (log);
\draw[darr] (engine) -- node[above,font=\footnotesize]{requests} node[below,font=\footnotesize]{stream} (prov);
\draw[darr] (prov) -- node[pos=0.5,bn]{1} (models);
\draw[darr] (engine.south east) -- node[pos=0.55,right=3pt,font=\footnotesize]{calls, results} (perm.north west);
\draw[darr] (perm) -- node[pos=0.5,bn]{2} node[above=5pt,font=\footnotesize]{run} (tools);
\draw[arr] (proj.south west) -- node[pos=0.45,bn]{3} (engine.north east);
\draw[arr] (ext) -- node[pos=0.5,bn]{4} (engine);
\draw[arr] (others.north) -- node[pos=0.5,bn]{5} (others.north |- host.south);
\draw[arr] (sched.north) -- node[pos=0.5,bn]{6} (sched.north |- host.south);
\end{tikzpicture}
\caption{Parts of a harness. Shaded parts form the \emph{trusted base}; the model and everything it reads are outside it. Numbered marks are trust boundaries: \bnd{1} model output, \bnd{2} tool and web content, \bnd{3} project files, \bnd{4} \emph{extensions}, \bnd{5} other agents' text, \bnd{6} turns that start with nobody there.}
\label{fig:parts}
\end{figure}
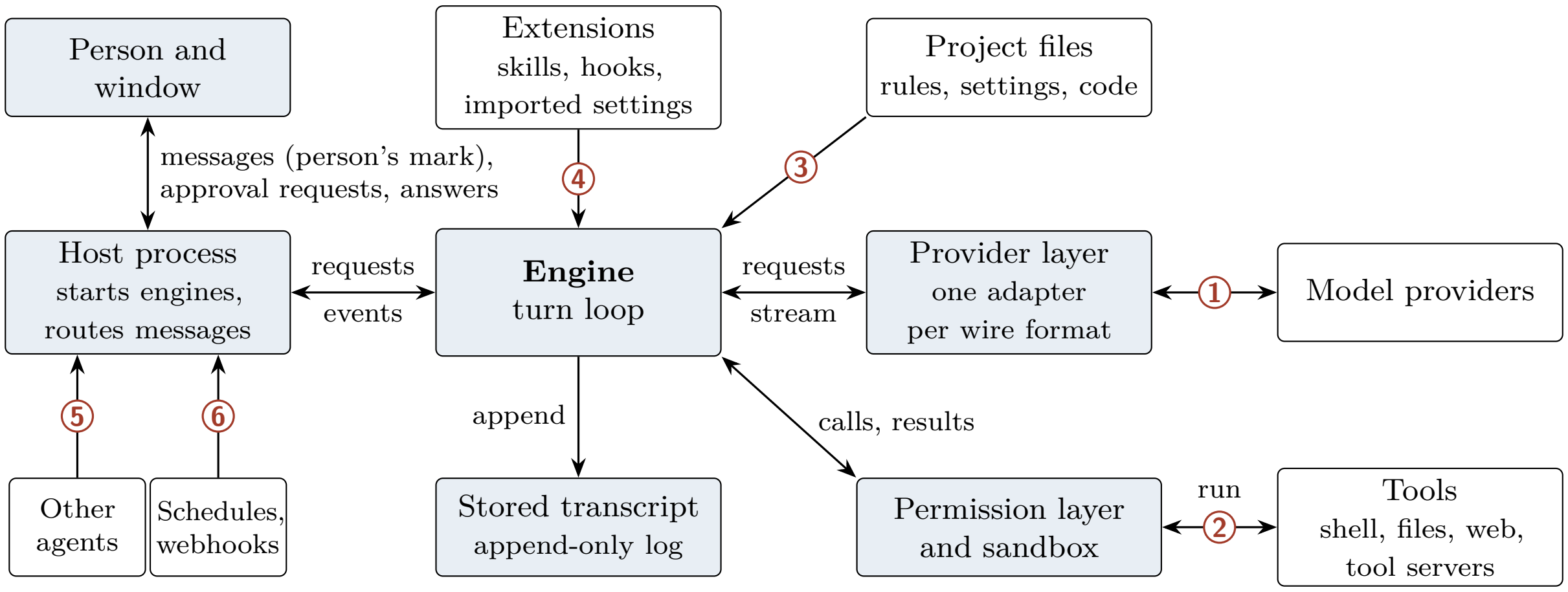

This section maps the parts of a harness before the decisions are listed. Most hard problems sit where something crosses from a part the harness controls into one it does not: the model's reply, a fetched web page, a file someone else wrote. Figure~\ref{fig:parts} shows the parts and those crossings.

\paragraph{Parts.} The \emph{person and window} are the source of authority: messages the person types carry a mark that says so, and approval requests are answered there. The \emph{host} starts and stops engines and routes messages, approval requests and answers. The \emph{engine} runs the turn loop: it assembles requests, reads streamed replies, decides what each means, starts tools, handles failures and records everything. The \emph{provider layer} has one adapter per wire format. \emph{Tools} run as child processes or, for the harness's own file tools, inside the engine. The \emph{permission layer} (rules, \gl{mode}, checker, approval requests) decides whether a call may run; the \emph{sandbox}, an operating-system \emph{profile} applied to each command, limits what it reaches. \emph{Extensions} (\gl{skills}, \gl{hooks}, \gl{tool servers}, imported settings) are files that can name programs or allow tools. The \emph{transcript} is an append-only log per chat from which everything else can be rebuilt. \emph{Other agents and schedules} send text or start turns. 

\paragraph{Not every harness has every part.} A single-process command-line harness folds host and engine into one process: the cards on \emph{transports} and hosting (S4, S5) and the parts of other cards about an engine stopped when idle and started again do not arise, while the cards on the turn, failures, permission and state apply unchanged. A harness without a sandbox skips P3 and P4; one that does not ask before unlisted commands skips P2. This mapping is reasoned (\Bel).

\paragraph{Trust boundaries.} The model is treated as untrusted: its reply is a request, not a decision. Six kinds of input cross into the harness and are data until the permission layer decides otherwise: \bnd{1} model output; \bnd{2} tool and web content; \bnd{3} project files, which teammates or whoever sent the repository may have written; \bnd{4} extensions; \bnd{5} other agents' text; and \bnd{6} anything that starts a turn when nobody is there. Cards behind each: \bnd{1} T1--T5, F4, F5, P2, P3, P7, P8; \bnd{2} T5, P9; \bnd{3} P1, P4, P6; \bnd{4} P6; \bnd{5} P9, S3; \bnd{6} P5. A systematisation of the agentic attack surface maps trust boundaries of the same kind~\cite{sok}. 

\section{The design space}\label{sec:space}

This section is the body of the paper: 30 decisions a harness makes, written down or not, grouped by the part that makes them. Each heading says whether the card applies to every harness or only to some. Table~\ref{tab:cards} lists them with their invariants in short form. For a builder starting out, we suggest ten cards first: T1, T2, T5, F1, F2, S1, S2, P1, P2 and V1 (the author's judgment, not a measurement). Section~\ref{sec:compare} summarises how ten other harnesses decide.

\begin{table}[!tp]
\centering\footnotesize
\renewcommand{\arraystretch}{1.05}
\caption{The 30 decision cards: the core clause of each invariant in short form; where the invariant was checked on our harness (script: a scripted provider; screen: on screen in the installed application with a scripted provider; real: a real model behind a fault relay; shell: commands run inside the real sandbox); the tag of the core clause; and the verdict on the matching checklist item for our harness under the scoring rule of Section~\ref{sec:cross}: H holds, N does not, D holds in a different form, U not tested. An item fails if any tested part fails, so a Tested core clause can stand beside an N, and an item can hold while a clause the checklist has no part for is Open. These verdicts are a self-assessment under the conditions of Section~\ref{sec:cross}.}\label{tab:cards}
\setlength{\tabcolsep}{4pt}
\begin{tabularx}{\linewidth}{@{}l>{\raggedright\arraybackslash}X>{\raggedright\arraybackslash}X l l c@{}}
\toprule
\textbf{Card} & \textbf{Decision} & \textbf{Core clause, short} & \textbf{Checked by} & \textbf{Tag} & \textbf{Verdict}\\
\midrule
\cd{T1} & How a turn ends & One recorded end reason per turn & script & Tested & N\\
\cd{T2} & Every tool call ends exactly once & Each started call shown once as ended & script & Tested & H\\
\cd{T3} & Early start, discarded attempts & Only read-only calls start early & script & Tested & N\\
\cd{T4} & Arrival order, reasoning, what is shown & Streamed text stays, once, in order & script & Tested & H\\
\cd{T5} & Tool results and long-running tools & Results non-empty, bounded, kept & script & Tested & N\\
\addlinespace
\cd{F1} & Retry budgets that do not multiply & Each failure episode bounded & script, real & Tested & N\\
\cd{F2} & Deadlines, silence, the ``gave up'' line & Silent provider ends by a deadline & script, screen, real & Tested & N\\
\cd{F3} & Limits, reset times, backup models & Only stated reset times shown & script, real & Tested & H\\
\cd{F4} & Guards for weak models; prompt fixes & Every guard bounded & script, screen & Tested &H\\
\cd{F5} & Refusals and content filters & Refused reply never shown as answer & script & Tested & N\\
\addlinespace
\cd{C1} & Assembling the prompt per family & Every prompt claim true for the session & script & \textbf{Open} & N\\
\cd{C2} & Keeping the cached prefix stable & Prefix bytes fixed from turn to turn & script & Tested & H\\
\cd{C3} & What must survive a summary & Rules come back from their source & script & Tested & U\\
\addlinespace
\cd{S1} & Source of truth, resume after a crash & Side effects logged before they run & script & Tested & H\\
\cd{S2} & A stop that is true everywhere & After stop no new tool or request & script & Tested & N\\
\cd{S3} & Queued and steered messages & Each message delivered once & script & Tested & U\\
\cd{S4} & Keeping transports identical & Same scenario, same behaviour per transport & script & \textbf{Believed} & N\\
\cd{S5} & Hosting the engine, the link to it & Stalled reader loses nothing & script & Tested & U\\
\cd{S6} & Cost booking and the budget stop & Each billed request counted once & script & Tested & H\\
\addlinespace
\cd{P1} & Modes, rule sources, precedence & Refusals decided first & script & \textbf{Open} & N\\
\cd{P2} & Reading a shell command exactly & Approve only exact shell readings & script & \textbf{Open} & N\\
\cd{P3} & What a sandbox can express & Deny renames and parents too & shell & \textbf{Believed} & U\\
\cd{P4} & Hiding credentials & Login places hidden from every tool & script & \textbf{Open} & N\\
\cd{P5} & Who is watching, per turn & Watched only with the person's mark & script & \textbf{Believed} & H\\
\cd{P6} & Extensions that grant themselves rights & No grants before trust & script & Tested & H\\
\cd{P7} & Approval requests & Harness reason shown first & script & Tested & U\\
\cd{P8} & Standing grants and refusal text & Grants out of the model's reach & script, shell & Tested & H\\
\cd{P9} & Other agents' text and outside content & Outside text fenced as data & script & \textbf{Open} & N\\
\addlinespace
\cd{V1} & One notice per event & One notice and one title per event & script & Tested & N\\
\cd{V2} & Helper agents & Helper status shows its outcome & script & Tested & H\\
\bottomrule
\end{tabularx}
\end{table}

\needspace{12\baselineskip}
\subsection{The turn}\label{sec:turn}

A turn is the unit of work a person asks for. The hard part is the edges: knowing it is over, tracking tools that run side by side or for a long time, discarding a half reply, and keeping the screen in the order things arrived.

\card{T1}{How a turn ends}{\TagT}{every harness}
\plain{A turn can stop for many reasons; the person must be told which, and see exactly one final answer.}
\begin{fields}
\item[Decision.] How does a turn end, and what does the person read? Finish, Stop, the \gl{step limit} (about 100 steps in our harness), a spend cap, a hook, a final failure, a safety stop or an output cap. If the end is not named, a turn that finished and a turn that died look the same. 
\item[Options.] 
\begin{opts}
\item[Where ends are decided:] scattered early returns (cheap to write; nobody can list them); one ``end with reason R'' call that writes both the record and the screen (auditable; every exit must use it); or a pure table from facts about the reply to one action (testable row by row; the facts must first be gathered in one place).
\item[How the closing is written:] plain text with no call ends the turn (simple; no end-of-turn check can run); or a \gl{finishing tool} asks the model for its closing (checks such as pending background work run first; the model may put words beside the call, and a common approach never ends the turn at that call).
\end{opts}
\item[Invariant.] (\textup{a}) \inv{every turn ends with exactly one recorded end reason}~\Tst{}; (\textup{b}) \inv{every end other than a normal finish says why on screen}~\Tst{}; (\textup{c}) \inv{exactly one closing message is stored and drawn}~\Tst{}. 
\item[Corner cases.] An error handled locally is stored as ``completed'', so the history looks healthy. An exit nobody listed (an abort inside a hook, an exception thrown out of the loop) skips the hooks and the end marker. A lead-in sentence beside the finishing call is taken as the closing, and the work stops early. The model rewords its closing when asked again, and two closings are drawn. A stream closes without a stop reason and is shown and stored as whole (seen in our harness, Section~\ref{sec:cross}), or the half-written call of a cut reply runs. A continue path skips the step counter, so the turn never reaches its limit. 
\item[Test.] List the exits with a script over the syntax tree; for each, one scripted scenario asserting one end record, its reason, the exit code of a \gl{one-shot run}, and that closing hooks ran. Script follow-ups to a finishing call (empty, identical, reworded, a lead-in with work pending) and streams cut after text, mid-call and after a \gl{heartbeat}; assert one closing, no half call run, and the step counter advanced on every continue path. 
\end{fields}

\card{T2}{Every tool call ends exactly once}{\TagT}{every harness}
\plain{Every tool shown as started must later show as finished, failed or not run, with one answer each.}
\begin{fields}
\item[Decision.] How is a result attached to its call, and what closes a call that never ends normally? Calls run in parallel, finish out of order, and may be discarded, cut, refused, stopped or never started. 
\item[Options.] 
\begin{opts}
\item[Matching:] a first-in-first-out list of open items (simple; when the second call finishes first, the wrong \gl{row} closes) or a map by call id with one terminal state per call (correct; every abnormal end needs a stand-in result). Table~\ref{tab:endings} lists 13 classes of ending; our harness's code has 21 distinct outcomes that fall into them.
\item[Storage order:] completion order (no waiting; a replay differs from run to run) or call order, each result written as soon as it finishes (replays are stable; a crash loses nothing that finished).
\end{opts}
\item[Invariant.] (\textup{a}) \inv{every call the person saw start is shown once as finished, failed or not run}~\Tst{}; (\textup{b}) \inv{every call the model issued has exactly one result, matched by its id and stored in call order}~\Tst{}. 
\item[Corner cases.] Calls finish out of order. The list of open items is not cleared between turns, so every later end closes the wrong row. A turn ended by its budget or by Stop leaves calls that never started without a result. A retried reply reuses the call ids of the abandoned attempt, so one id gets two results. A weak model sends blank or duplicate ids in one reply. 
\item[Test.] One scripted scenario per terminal state; after each turn assert zero open items and as many results as calls. Two parallel calls with the second finishing first. Stop mid-reply, then run a second turn and check the ids. Kill the process after the second of three results; assert the first two are present, in order. 
\end{fields}

\card{T3}{Early start of read-only calls, discarded attempts}{\TagT}{every harness}
\plain{Simple reads may start early, but a reply that is thrown away must leave nothing behind.}
\begin{fields}
\item[Decision.] May a call start before the reply ends, and what happens to the calls of an attempt that is thrown away? Waiting costs seconds per read, yet a reply can still stall, drop, be refused or be retried. 
\item[Options.] 
\begin{opts}
\item[When tools start:] only after the reply ends (simple; seconds lost on every read); or read-only calls start as soon as their arguments are complete, while writes and anything needing approval wait (a common approach; fast, but a retried reply reruns its reads, a waste the harnesses tested accept); or early results are kept for the retry (seen in no harness, \Opn).
\item[A discarded attempt:] ignored (simple; its results are orphaned and its half text stays beside the full answer); or its early calls are stopped and given a ``discarded'' result, and its half text is withdrawn from every \emph{output mode}: window, one-shot text, streamed records (nothing stale remains; the retry runs those calls again; our harness); or the partial text is kept, marked, and the model asked to continue from it (no text is paid for twice; the answer arrives in two pieces; seen in other harnesses after a reply cut at its length limit).
\end{opts}
\item[Invariant.] (\textup{a}) \inv{only read-only calls start before the reply ends}~\Tst{}; (\textup{b}) \inv{a cut or abandoned attempt is either withdrawn, leaving no text, result, open item or running process behind, or kept, marked as partial and continued; it is never shown or stored as complete, and none of its calls runs unknown to the model}~\Opn{} (a stream that closed with no finish reason was shown and stored as complete in our harness, Section~\ref{sec:cross}; the clause is tested only for attempts the harness itself recognises as cut). 
\item[Corner cases.] A reply is refused after its early read ran, and the model is told ``not run'' for a read that did run. One output mode prints the abandoned attempt beside the kept one. A counter counts attempts instead of kept replies. A command started by a dropped reply keeps running under an id the model never learned. 
\item[Test.] Drop the stream mid-reply with an early read in flight; in each output mode assert one tool row, no orphan result, no leftover process, one answer and unique ids. Read the result text in the next request after a scripted refusal. Count executions after a retry. 
\end{fields}

\card{T4}{Arrival order, reasoning, and what is shown}{\TagT}{every harness}
\plain{What the person saw must stay, once and in order, and private reasoning must not leak into the answer.}
\begin{fields}
\item[Decision.] In what order are a reply's parts stored and drawn, and where does reasoning go? Item-based wire formats may close the opening message after the call; reasoning comes in fields, inline tags or signed blobs some models need back. 
\item[Options.] 
\begin{opts}
\item[Stored order:] as items close (simplest; the opening sentence lands after the call) or as items arrive, a closed item waiting until every earlier item has closed, using the provider's output index or first-seen order (matches what the person saw stream; needs that wait).
\item[Display of steps:] one group row with all text inside (compact; the model's narration is hidden) or each text its own row with the steps between two texts grouped (nothing folded away; more rows; common in several harnesses, \Bel).
\item[Reasoning:] shown raw (it leaks into the reply); inline tags moved into a reasoning block (clean; one parser per tag style); sent back to models that require it (needed by them; more tokens); signed reasoning from another conversation stripped once and the strip recorded (works after a model switch; must not repeat on resume).
\end{opts}
\item[Invariant.] (\textup{a}) \inv{every text the person saw while streaming is stored after the turn once, in arrival order, and is on screen after the turn}~\Tst{}; (\textup{b}) \inv{a repeated text is drawn only once}~\Bel{}; (\textup{c}) \inv{no reasoning appears inside the visible reply}~\Tst{}; (\textup{d}) \inv{a model that requires its reasoning back receives it}~\Tst{}. 
\item[Corner cases.] The opening sentence is drawn as a folded reasoning row. The first reply of a chat vanishes into a step group when the next turn ends. The same text is stored twice and drawn twice. Inline reasoning tags stay visible in the reply. A provider rejects signed reasoning after a model switch. 
\item[Test.] Script events in the provider's real order (text start, call start, call end, text end); assert stored order [text, call]. Render a recorded event trace to rows and assert the finished rows are the ones that were live. For each reasoning shape, assert the visible text, the stored blocks and what the next request carries. 
\end{fields}

\card{T5}{Tool results and long-running tools}{\TagT}{every harness}
\plain{Tools may return nothing, too much, or run for ten minutes; the model still needs a short, true answer.}
\begin{fields}
\item[Decision.] What does the model get from a tool whose output is empty, huge or unchanged, and what happens to a command that outlives its timeout? Huge output floods the context; empty output looks like failure; a blocked model can do nothing else. 
\item[Options.] 
\begin{opts}
\item[Large output:] truncate (simple; the rest is lost) or save big output to a file and send a preview and the path (nothing is lost; the model must fetch more itself).
\item[Empty or repeated output:] an empty result becomes a sentence such as ``(completed with no output)'' (never looks like a failure); an unchanged re-read returns ``unchanged'' (saves tokens; must still count as a repeat for the guards of F4).
\item[Long commands:] kill at a fixed timeout (simple; work is lost) or move a command past its timeout to the background, tracked in a list of background commands that sends a finish notice and kills the process group on Stop (work is kept; the harness must maintain that list).
\item[Background work in a watched turn:] hold the turn open (the person sees ``working'' for an hour and cannot tell why) or end the turn and start a new one when the finish notice arrives (a common approach, \Bel).
\item[Old results near the context limit:] blank all but the last few (frees room; the evidence is gone) or keep one line each, such as ``ran X, exit 0, 40 lines'' (keeps a trace; \Bel).
\end{opts}
\item[Invariant.] (\textup{a}) \inv{every tool result the model sees is non-empty and bounded while the full output stays retrievable}~\Tst{}; (\textup{b}) \inv{every call returns to the model within its timeout; a command past its timeout is either killed with a result that says so, or moved to the background and tracked until its end is reported}~\Opn{} (in our harness's one-shot mode a command with a 3\,s timeout was reported as timed out, yet ran on, finished its side effect, and held the turn for 45\,s; Section~\ref{sec:cross}). 
\item[Corner cases.] A result says a command timed out while the command runs on. Secrets in a saved output are not redacted. An ``unchanged'' note is not counted as the earlier result, so the repeat check fires late. A stop must kill this turn's processes but not earlier background work. Clearing removes results that carry decisions (answers, helper reports, plans), or it happens without telling anyone. 
\item[Test.] Tools returning empty, huge and secret-bearing output: assert the cap, the saved file and the text sent. A command sleeping past its timeout: assert the backgrounded result, a later finish notice, and that Stop kills its process group. A history with every result kind: trigger clearing and assert the kept set and one plain notice. 
\end{fields}

\subsection{When things fail}\label{sec:fail}

Providers go silent, drop connections, run out of quota and refuse; weaker models send empty replies or repeat themselves. The common thread of these cards: every wait has a limit somebody chose, and every line shown about a failure is true.

\card{F1}{Retry budgets that do not multiply}{\TagT}{every harness}
\plain{Retrying is fine, but the total wait needs one limit that someone chose and the person is told.}
\begin{fields}
\item[Decision.] Who decides whether a failure retries, waits, switches or ends the turn, and for how long? Three places see a failure: the code that sends the request, the turn loop, and the branch that handles a stalled reply. If each retries on its own, their retry counts multiply. A \emph{failure episode} runs from the first failure until an answer arrives. 
\item[Options.] 
\begin{opts}
\item[Who decides:] each place where a failure is seen (simple; request retries times loop retries times \gl{waiting rounds} reach tens of minutes); one pure function from the classified error, the model family, what was already shown, the counters, the time elapsed and whether anyone is watching, to one action: compact and resend, a smaller output cap, the next account, the next \gl{backup model}, retry after a delay, one extra send, wait, or fail with this line (testable as a table; many inputs); or rules as data with a list of errors never retried (readable; the counter resets live elsewhere).
\item[What retries:] our harness never retries authentication errors or bad requests, backs off with jitter from 0.5\,s to about 32\,s, and treats a \gl{retry-after} longer than about a minute as a limit, not a retry (F3). Table~\ref{tab:errors} gives all 23 rows from failure to action.
\end{opts}
\item[Invariant.] (\textup{a}) \inv{the total time and attempts spent on one failure episode are bounded by stated numbers, and the first one reached ends it}~\Tst{}; (\textup{b}) \inv{no layer retries a failure another layer has already retried}~\Opn{} (in our harness the request layer tries three times inside each of six rounds separated by five waiting rounds, Section~\ref{sec:cross}). 
\item[Corner cases.] Each waiting round restarts the inner retry count; our harness does this and relies on the five-waiting-round cap and the episode deadline to cap the total, without which the budgets multiply. Two bounds on one episode can disagree with what the person is told: behind the relay the waiting-round cap ended an overload episode at about 260\,s while the waiting row named the ten-minute deadline (F2). A stall branch decides outside the function, where its tests never look. Which counters should reset after an account switch is \Opn. 
\item[Test.] A table test of the deciding function, one row per error, family and counter state. A scripted provider per failure shape through the whole loop. A \emph{fake clock} and a provider that never answers: assert the end at the deadline and count the requests sent. Enumerate failure kinds and assert each reaches the function. 
\end{fields}

\card{F2}{Deadlines, silence and the ``gave up'' line}{\TagT}{every harness}
\plain{A model that is thinking and a provider that is dead look the same on screen; say what the harness waits for and when it gives up.}
\begin{fields}
\item[Decision.] How long to wait for a silent or failing provider, what to show meanwhile, and what the last line says? 
\item[Options.] 
\begin{opts}
\item[Detecting silence:] one idle timeout (simple; misfires on models that reason silently) or separate time limits before the first byte and between chunks, longer for slow thinkers, with heartbeats counting as life (correct; needs a table per family, and a silent connection may wait 3 to 5 minutes before its first reconnect). A short detector of about 30\,s, used by some harnesses, suits only streams without silent reasoning.
\item[Deadline:] none (a dead provider keeps the chat ``working'' for ever) or one counted from the first failure and cleared by an answer, about ten minutes when watched and an hour or more when unwatched (bounded; may give up on a slow recovery). Our harness also caps waiting rounds at five; whichever bound comes first ends the episode, so the deadline shown must be the earlier of the two.
\item[After text was shown:] keep retrying through the long waiting rounds (more chances; a retried answer can repeat or contradict what the person already read) or, once the person has seen part of the reply, no waiting rounds: one more send after the deadline, then a final line (nothing read is contradicted; such a turn gives up sooner).
\item[\emph{Waiting line}:] a spinner in the window (no engine work; blind to hidden reasoning and to the reconnect time) or, after a fixed silence counted from sending (20\,s in our harness; elsewhere a colour change at 10\,s or a line at 60\,s), one plain line from the component that sees the stream: what it waits for, how long, when it reconnects, and that Stop cancels (honest; needed on every route). The time before a reconnect may differ by route and by where the provider runs; the waiting line's own threshold need not.
\end{opts}
\item[Invariant.] (\textup{a}) \inv{a chat whose provider never answers ends by a deadline}~\Tst{}; (\textup{b}) \inv{the stop time shown in the waiting row is the time it really stops}~\Opn{} (false: under overload on every request the row named a stop time five to six minutes after the real end, because the waiting-round cap ends the episode first; Section~\ref{sec:realruns}); (\textup{c}) \inv{its final line is plain, names the next step and says the message and work are saved}~\Tst{}; (\textup{d}) \inv{these hold for chats driven through an \gl{outside agent program}}~\Opn{}. 
\item[Corner cases.] A route shows no waiting line, or shows it only after the stream opens (our harness's one-shot output showed nothing for the first 300\,s of a silent provider, Section~\ref{sec:cross}). A countdown to the reconnect rises between two updates of the waiting line (seen in real runs). Chats in which the harness drives an outside agent program have no silence clock and no deadline; which deadline they need is \Opn. Time blocked on a slow reader is counted as silence. A row says ``answered after 5 tries'' when no try answered (\Bel). Ten retry rows stack up instead of one row updated in place. The machine's sleep eats the budget. The final line uses jargon such as ``stream'' or ``watchdog''. 
\item[Test.] For each route that has a deadline, a fake clock and a silent provider: assert the waiting line at the threshold with the true remaining time, its removal when output arrives, one retry row, ``gave up'' at the deadline, and a final line that passes a forbidden-word scan and names an action. Inject a fault after text has been shown: assert no waiting rounds, one more send, and a final line. 
\end{fields}

\card{F3}{Limits, reset times and backup models}{\TagT}{harnesses with several accounts or models}
\plain{Switching models is fine, but never invent a reset time, and always say which model answered.}
\begin{fields}
\item[Decision.] When a model or account is out: wait, switch account, switch model, or stop, and say what? Overload, a short rate limit and a spent usage limit can arrive with the same status code. The first needs a wait of seconds, the second about a minute, the third a stop until the reset time the provider states. 
\item[Options.] 
\begin{opts}
\item[Classifying:] retry everything (simple; a spent weekly limit is retried for minutes) or sort by wording into usage limit (stop and give the stated reset), short rate limit (retry) and overload (say the service is busy) (the right action each time; wording lists are fragile).
\item[On a spent account:] the next account of the same provider before a backup model (same model, same request; needs several accounts) or straight to a backup model (works with one account; changes the model).
\item[What triggers a backup:] overload and unavailability only, a common approach (fewer surprises; a limit stops the chat), or also server errors and limits, through a chain of at most about three models that keeps call--result pairs and drops the first family's private reasoning (keeps working; another vendor may answer).
\item[Telling the person:] logs only (quiet; the person never knows), a row at the switch (some harnesses; later replies carry no mark), or also a label on each reply a backup wrote (always visible).
\item[Return:] after a switch for one request, back at the next turn without a row (no noise; only the reply label shows the switch); after a switch held until the reset, one ``back on'' row, exactly once (the person knows; the return must wait for the account's own reset, not a timer). One approach announces every return.
\end{opts}
\item[Invariant.] (\textup{a}) \inv{every reset time shown was stated by the provider}~\Tst{}; (\textup{b}) \inv{every reply written by a model other than the one the person picked is labelled}~\Tst{}; (\textup{c}) \inv{the chat returns to that model only when its account is free}~\Tst{}. 
\item[Corner cases.] ``Weekly limit, try again in two days'' matches a rate-limit phrase and is retried. A guessed pause is printed as ``resets at''. The \gl{backup hold} expires before the \gl{spent-account record}, so ``back on'' is followed by a failure. A provider with a single key records nothing, so every chat fails on it again. A limit hit at turn start is thrown outside the turn and kills the engine. A short rate limit is retried correctly but its row uses the overload wording (``is busy''), seen in real runs. Whether silence should trigger a switch is \Opn. 
\item[Test.] (1) Recorded real error bodies per kind, with expected class and text. (2) An error with no reset: the text names no time. (3) Two accounts, the first spent: the second serves the same request. (4) Main and backup on different wire formats: the backup's request shape, the switch row, the label, the return row. (5) The backup hold expires before the spent-account record: the chat stays on the backup and no turn fails. 
\end{fields}

\card{F4}{Guards for weak models, and what a prompt cannot fix}{\TagT}{every harness}
\plain{Weak models may be nudged a few times, visibly; a sentence in the prompt is not a reliable fix.}
\begin{fields}
\item[Decision.] Which bad reply shapes are re-asked or repaired, how often and how visibly, and when is a prompt sentence enough? Weak models send empty replies, announce work without a call, repeat calls, loop, or call missing tools; providers reject fields such as an effort level. 
\item[Options.] 
\begin{opts}
\item[Bad reply shapes:] end the turn on any of them (honest; unusable on weak models) or bounded \gl{weak-model guards} (works; each guard needs a signal, and word lists are fragile): re-ask an empty reply at most twice with visible waits (our harness does this for most families; for one family a coded rule instead ends the turn at once with a plain line when the first reply is empty, and nudges once when an empty reply follows tool results); nudge an announcement without a call at most three times; warn, then stop, a repeated call; cut a repeating stream once.
\item[Repairs:] repair a malformed call only when exactly one reading exists (saves the turn; a wider repair could change what the call means); repair a request field the provider rejects, such as an effort level, once per kind and remember it for the \gl{session} (one failure per kind; a second failure must be shown).
\item[A second model as judge:] (more cover; less predictable).
\item[A prompt sentence:] (cheapest; it changes how likely a habit is, not whether it happens; \Bel, case~4).
\end{opts}
\item[Invariant.] (\textup{a}) \inv{every guard acts at most a stated number of times}~\Tst{}; (\textup{b}) \inv{each re-ask, nudge, repeat stop and cut is shown to the person with its real count}~\Tst{}; (\textup{c}) \inv{a repair never changes what a call means}~\Tst{}. 
\item[Corner cases.] The row says ``in 1\,s'' for a 5\,s wait, or ``5 times'' after six calls. The give-up line after counted re-asks gives no count (our harness, Section~\ref{sec:cross}). An ``unchanged file'' note is not counted as a repeat. The announcement nudge runs only for helpers, not in a normal chat. Word lists miss other languages; a better signal is \Opn. Reserved object keys in nested arguments become real keys. A ``do not repeat'' prompt sentence leaves duplicate closings as frequent as before. A model claims a check it never ran, and no harness read catches this at runtime (seen once, \Bel). 
\item[Test.] Script empty replies one to three times, announcements in several phrasings and languages, repeated calls, a repeating stream, a table of malformed calls, and a provider that rejects each request field once; assert counts, rows and final lines as strings. For any prompt-wording fix, count the habit in ten real runs on the affected family before and after. 
\end{fields}

\card{F5}{Refusals and content filters}{\TagT}{every harness}
\plain{A refused reply is never shown as an answer, and nothing it asked for may run.}
\begin{fields}
\item[Decision.] What happens to a reply stopped by a safety check or a content filter, to its calls, and to the line shown? It may hold text and calls, and an early read may already have run. 
\item[Options.] 
\begin{opts}
\item[Refused reply:] show the provider's raw text (simple; misleading) or keep the refused reply's text, run none of its calls, and ask once to continue without that part (true and recoverable; more rows to word).
\item[Content filter:] ask again up to five times with backoff and withdraw blocked text that had no call (bounded; every ask is a visible row).
\item[Explaining it:] the provider's own reason behind a control that opens details, and a policy link only for the family whose policy it is (names the true owner; one link per family to keep).
\end{opts}
\item[Invariant.] (\textup{a}) \inv{a blocked or refused reply is never shown as an answer, and none of its unstarted calls run}~\Tst{}; (\textup{b}) \inv{the line names the true reason and its owner}~\Opn{} (on our harness's generic route a refusal field was announced as an empty response and re-asked; Section~\ref{sec:cross}); (\textup{c}) \inv{the model is told which calls did run}~\Tst{}. 
\item[Corner cases.] A refusal links another vendor's policy. Blocked text piles up on screen. A filter retry happens without a row. Early reads that ran are reported as ``not run''. 
\item[Test.] Script a refusal once and a filter five and six times, each with and without calls; assert the rows, the withdrawn text, the final line, the link's owner, and the result text the next request carries. 
\end{fields}

\subsection{What the model is told}\label{sec:told}

Every request starts with text the harness writes: who the model is, what it may do, which tools and rules exist. The model acts on all of it; it is paid for on every request and cached only while its beginning, the \emph{cached prefix}, stays byte-identical; and after a summary some of it must come back from its source.

\card{C1}{Assembling the prompt per model family}{\TagO}{every harness}
\plain{Everything the prompt says about the session must be true, because the model acts on it.}
\begin{fields}
\item[Decision.] What is each model told about itself, its environment and its tools, and does it differ by family? Families were trained on different prompts and act on every sentence. 
\item[Options.] 
\begin{opts}
\item[Prompt per family:] one prompt for all (one text and one cache; pulls families toward habits not their own, and costs tokens); a prompt per family modelled on that family's own harness (smallest, plausibly the best fit, \Bel; more text to maintain); or a shared core plus a short family block (a middle path).
\item[Environment sentences:] copied from elsewhere (cheap; false wherever the setting differs) or emitted only when this session's settings make them true (correct; one more input to the prompt).
\item[Helpers' role text:] chosen by a default argument (no work; a helper can get the role of another mode) or stated explicitly wherever a helper is started and checked in the helper's request (the role always fits; one more argument at every call site).
\end{opts}
\item[Invariant.] (\textup{a}) \inv{every factual claim in the assembled prompt (identity, sandbox, network, branches, reviewers, tool names) is true for the session that receives it}~\Opn{} (our harness's prompt told the model to use two tools absent from the request's tool list; both can be loaded on demand, and whether the model could load them was not checked; Section~\ref{sec:cross}); (\textup{b}) \inv{each behaviour rule is stated in exactly one section}~\Tst{}. 
\item[Corner cases.] A chat is told that a reviewer will merge its branch when no such machinery runs. The prompt says ``not a sandbox'' while a tool text says commands are sandboxed, so ``operation not permitted'' reads as a broken machine. ``Hold off on tests'' stands beside ``run the tests''. A checklist names a tool the session lacks. A ``do the work now'' nudge acts on a question. One person's routing rule is shown to every family. A helper started in one mode receives another mode's role prompt (``a worker reporting to a lead'') through a default argument. A model id behind a \gl{gateway} selects the wrong family. 
\item[Test.] Build the real prompt for every family and chat kind; assert forbidden phrases are absent, the identity sentence appears once, no pair from a list of mutually exclusive claims appears together, and every tool-like token is in the same request's tool list. Run a two-helper task on a scripted model and assert the helpers' request text holds no role their mode lacks. Table-test family selection over a list of ids, gateway-prefixed ids included. (These two tests are planned, not run.) 
\end{fields}

\card{C2}{Keeping the cached prefix stable}{\TagT}{harnesses on providers with prompt caching}
\plain{Anything that changes early in a request quietly makes every later turn cost more.}
\begin{fields}
\item[Decision.] What starts every request, in what order, and what may change there? Caches match from the first byte, so one unstable early value bills the whole conversation again, unseen by the person. 
\item[Options.] 
\begin{opts}
\item[Order:] tools, fixed prompt, per-chat role, then first-message notes, most stable first, with changes appended (cache-friendly; changes reach the model late); everything in the system prompt (simple; any change breaks the cache); or the instruction block rebuilt on each request (no summary can drop it; it must stay byte-stable).
\item[Volatile values] (models connected, quota left): in tool text (always current; each change, and each engine restart after one, bills the whole conversation again); in a first-message section, with ``changed since last told'' reminders appended (stable prefix; the model learns of a change one message later); or behind a list tool (nothing in the prefix; one more call when the model needs it).
\item[\emph{Posture}:] coding or general, decided once per chat from the working folder and then fixed (a smaller prompt for chats that are not about code; a folder that changes later is ignored; \Bel).
\item[Size ceiling:] a test per family that fails when the fixed text grows past a written ceiling (growth is noticed; raising the ceiling needs a stated reason; \Bel: planned).
\item[Cache lifetime:] where a provider offers a short and a long cache lifetime, the harness asks for one with each request. Short is cheaper to write, but a person who pauses longer pays for the whole chat again; long costs more to write but survives breaks. Some harnesses pick long while a person is typing and short for machine-paced work (\Bel).
\end{opts}
\item[Invariant.] (\textup{a}) \inv{bytes before the first appended message depend only on inputs that cannot change within a chat, so the tool list and system prompt are byte-identical from one turn to the next (tested on one vendor's direct route)}~\Tst{}; (\textup{b}) \inv{they stay byte-identical across engine restarts}~\Bel{}. 
\item[Corner cases.] A ``models connected now'' list in a tool description is cached again at every engine restart. A change of tool order between requests breaks the cache; sorting tools by name prevents it. A tool loaded mid-chat is appended to the tool list. A summary request is built unlike the main one and pays full price. A gateway route sends no cache markers. 
\item[Test.] Record request bodies with a scripted provider; replay a chat with each mid-chat change (a rules edit, a skill, a tool load, a new model list, a project file appearing in the working folder) and assert the prefix bytes are unchanged; restart with different connected models and compare tool text; assert each family's fixed text is under its ceiling after proving the test fails at 10\% less; with a real provider, check cached-read tokens on turn two. 
\end{fields}

\card{C3}{What must survive a summary}{\TagT}{every harness with long chats}
\plain{After a summary, the person's words and the measured facts come back verbatim, not paraphrased.}
\begin{fields}
\item[Decision.] When the chat is summarised, what must the next request still hold, and where does it come from? Summaries paraphrase; instructions, decisions, attachments and measured state are the hardest to recover. 
\item[Options.] 
\begin{opts}
\item[Trigger:] the context limit minus a reply reserve and a margin (uses the whole context; the summary must finish in time); a fixed share of the context limit, half in one harness read and 85\% in another (simple; early or late); or a summary built in the background (no wait; more machinery). A context-too-long error also triggers compact and resend, at most about three times, and the learned context limit is saved per model.
\item[Summary content:] free-form (cheapest; drops constraints); fixed headings with constraints quoted exactly (keeps constraints; the rest is still a paraphrase); an update of the last summary (avoids a copy of a copy; from comparison only, \Bel); or a harness checklist of measured facts with one repair request (catches omissions; one more call).
\item[The person's words:] summarised by the model (short; only the model's copy of the instructions survives) or kept verbatim up to a cap, oldest or largest dropped first (exact; one huge paste can fill the cap).
\item[Rules:] replayed from the first message (no extra work; after a summary they sit farthest from the newest text) or rebuilt from their files at each summary, restart or model switch (exact; one more read).
\end{opts}
\item[Invariant.] (\textup{a}) \inv{after any number of summaries the next request contains the standing rules taken from their source}~\Tst{}; (\textup{b}) \inv{it contains every typed message of the person that fits the cap, verbatim}~\Tst{}; (\textup{c}) \inv{it contains every fact the harness measured (changed files, open tasks, failing commands)}~\Tst{}. 
\item[Corner cases.] Found by \gl{probe} (\Bel): a decision given by clicking an option is a tool result, not typed text, and is dropped; attachments are reduced to ``[image]''; one huge first paste fills the whole cap; the changed-files list is rebuilt empty after a restart. Seen once (\Bel): instructions from a tool server that connected late are lost; two weak summaries in a row switch summarising off for good (one harness read retries after five minutes). 
\item[Test.] Apply a lossy scripted summariser three times; after each pass assert the rule block equals the file text, the typed messages are verbatim, and the measured facts are present. Restart after a summary and compare the changed-files list with its value before. Return ``too long'' four times and assert three resends and a final line. Assert one plain notice per compaction. 
\end{fields}

\subsection{State}\label{sec:state}

A chat must survive a crash, a Stop, a message arriving mid-turn, the same engine being run in two ways, and money being spent on attempts that failed. Each is a question about where the truth lives and what is written when. The recurring answer: an append-only log, written before anything is shown, with everything else rebuilt from it.

\card{S1}{The source of truth and resume after a crash}{\TagT}{every harness}
\plain{After a crash nothing finished is lost, and the model is told plainly what is uncertain.}
\begin{fields}
\item[Decision.] Where does a chat's durable truth live, when is it written, and what does resume do with a half-finished turn? After a crash, what is in memory, what is in the database and what was on screen can each tell a different story. 
\item[Options.] 
\begin{opts}
\item[Store:] an append-only log plus an index rebuilt from it (common; one truth); a relational table (queryable; replay state is lost on restart); or state spread over a record, a file and memory (what a harness drifts into without a plan; three versions after a crash).
\item[When to write:] at step boundaries: a call with side effects before its tool runs, each result before it is shown (nothing shown is missing after a crash; one write per step).
\item[How far to write:] hand each record to the operating system in one append (survives a killed process; our harness) or also force it to the device (survives a power loss; slower).
\item[One writer per log:] a lock file holding the process id and start time (simple; a stale lock must be taken over, which has races; our harness refuses a second engine after a short wait) or a \emph{lease} with an expiry checked at each write (no stale lock; a check on every write).
\item[Resume:] end a torn last line and skip unreadable lines on read (our harness), or cut the log back to its valid prefix (from comparison); close the open turn; give a cut call a stand-in result, or refuse new messages to the chat until the person acknowledges the lost turn (stricter; seen in other harnesses); optionally continue with a note and an attempt cap (\Bel) (nothing finished is lost; the model must be told what is uncertain).
\end{opts}
\item[Invariant.] (\textup{a}) \inv{a call with side effects is written to the log before its tool starts, and each result before it is shown}~\Tst{}, where ``written'' means handed to the operating system, not forced to the device; (\textup{b}) \inv{after any kill the log can be read, the open turn is closed, and either every call has exactly one result (real, or a stand-in saying ``outcome unknown, check before repeating'') or the chat refuses to continue until the person resolves the unknown outcome}~\Tst{}; (\textup{c}) \inv{a torn last line is ended and skipped on read, so no complete record is lost}~\Bel{}; (\textup{d}) \inv{a call that never started says ``not run''}~\Opn{} (true for calls dropped in a live turn; after a crash such a call still reads ``outcome unknown''). 
\item[Corner cases.] The last line of the log is torn. A call that never started is labelled ``unknown'' instead of ``not run''. A crash marker survives a deliberate stop, so a restart continues work the person stopped (\Bel). Two engines append to one log. The task list and the changed-files list are rebuilt empty from the history after a summary and an engine restart. Status is kept as timestamps the log cannot rebuild. A crash leaves a tool's own process running, so a call reported as unknown or interrupted completes later (seen in other harnesses; in ours the command died with the engine in 3 of 3 runs, but ran on after a Stop and after a timeout, Section~\ref{sec:cross}). 
\item[Test.] Kill the engine at fixed points of a scripted turn (before, during and after a tool; after the second of three results); restart and assert the invariant. Start two engines on one log, kill one, start a third. Delete the index and rebuild it from the log. 
\end{fields}

\card{S2}{A stop that is true everywhere}{\TagT}{every harness}
\plain{After Stop nothing new starts, open questions close, and the model later hears the truth.}
\begin{fields}
\item[Decision.] What does Stop mean for running tools, open approval requests, helpers, later requests, the model's next text and the process exit? A stop races tools, retries, approval requests and helpers. 
\item[Options.] 
\begin{opts}
\item[Delivery:] a keystroke or signal (no protocol; carries no reason) or a typed request, ahead of the queue, naming who stopped (reliable; needs a protocol).
\item[Checking:] once per step (cheap; a tool can slip through) or before every tool start and every new request (closes the race; more checks).
\item[Running tools:] kill the process group (clean; partial work is lost); background work may keep running if the model is told.
\item[What the model is told:] nothing (the model guesses) or distinct texts for ``nobody refused it'', ``did not run'', ``did not finish'' and ``stopped by another agent'' (true; more cases to word); two harnesses tested do the same (\Bel).
\end{opts}
\item[Invariant.] (\textup{a}) \inv{after a stop is accepted no tool starts and no model request is created}~\Tst{}; (\textup{b}) \inv{every open approval request of the turn is withdrawn}~\Tst{}; (\textup{c}) \inv{helpers of the chat stop and say who stopped them}~\Tst{}; (\textup{d}) \inv{the model is later told only what is true}~\Opn{}; (\textup{e}) \inv{the process exits with its stated code after running the session-end hooks}~\Opn{}. In our harness's one-shot mode, a Ctrl-C during a 30\,s command gave the result ``aborted before completion'' at once, but the command ran on and finished its side effect, and the process waited for it and exited 0 after about 30\,s (3 of 3; Section~\ref{sec:cross}); (d) and (e) are false there. 
\item[Corner cases.] A stop that closed an approval request is reported to the model as ``denied, do not retry''. A stop during a retry wait or a backup switch is handled as a failure. A late Allow on an approval request still on screen runs the step. The engine crashes on exit after a correct stop and skips the session-end hooks. A one-shot run killed between taking its lock and installing its signal handler leaves its command running and its lock behind. A test counts same-millisecond arrivals as ``after''. One stop test of our harness failed once in ten grouped runs with no cause found. 
\item[Test.] Hold the request open in a scripted provider. Stop at each phase (before send, mid-stream, while an approval request is open, during a slow hook, mid-tool, during a retry wait), and repeat the open-approval and slow-hook phases thirty times to catch races; count requests by sequence number, not timestamp; assert no new tool or request, no live child process, the class of each result text, the exit code, and that hooks ran. 
\end{fields}

\card{S3}{Queued and steered messages}{\TagT}{every harness}
\plain{A message typed during work arrives exactly once, in this turn or the next.}
\begin{fields}
\item[Decision.] What happens to a message sent while a turn runs? People and other agents keep typing. 
\item[Options.] 
\begin{opts}
\item[When it lands:] queued for the next turn (simple; the model learns late); joined to the running turn at the next step (to \emph{steer} it) and acknowledged (lands now; needs a step boundary); or refused while the turn is closing (no race; the person resends).
\item[Where the queue lives:] in memory (simple; lost on restart) or stored with the chat (survives a restart; one harness tested does this).
\end{opts}
A person's message never steers an unwatched turn; it starts a watched one (P5).
\item[Invariant.] (\textup{a}) \inv{each sent message is delivered in exactly one model request}~\Tst{}; (\textup{b}) \inv{it is shown in exactly one place}~\Bel{}; (\textup{c}) \inv{it survives an engine restart}~\Tst{}. 
\item[Corner cases.] A message arrives just as the turn closes. A late message is answered twice. A just-sent message is shown both in the queue and as a sent message in the chat (\Bel). Host and engine each keep a queue and other agents' messages arrive through a third, so one message is delivered twice unless the three are de-duplicated. A steer arrives during a compaction. 
\item[Test.] Send a message mid-stream, between steps and during the closing call; send three during a turn, kill the engine, and assert each was delivered once and in order; send during an unwatched turn and assert it starts a watched turn. 
\end{fields}

\card{S4}{Keeping transports identical}{\TagB}{harnesses run in more than one way}
\plain{If one engine runs in two ways, both ways must behave the same.}
\begin{fields}
\item[Decision.] How does a harness that runs one engine in more than one transport (interactive, one-shot, served to a window over a pipe) keep them from drifting apart? 
\item[Options.] 
\begin{opts}
\item[How drift is caught:] a recorded \emph{baseline} per transport, each checked on its own (cheap; a transport nobody runs drifts unnoticed) or a parity run that plays every scenario in every transport and compares requests, log lines, tool results, output and exit code (drift is caught; every intended difference must be named).
\end{opts}
\item[Invariant.] \inv{For every scripted scenario a transport can drive, each such transport produces the same requests, log lines, tool results and exit code, except for differences listed as intended}~\Bel{} (no fully passing cross-transport run is recorded; in our harness 41 of the 65 parity scenarios can run over the pipe, so the other 24 are checked in one transport only). 
\item[Corner cases.] One transport's baselines fail while the default run passes: in our harness a parity run of every scenario in two transports found 32 baselines and six test files of one transport failing unnoticed (case~3). Tool servers race the first request. A date normaliser runs in a different time zone from the engine. A parity run uses a mode no real chat uses. 
\item[Test.] In the automated checks, run every scenario in each transport and compare the five outputs above. When a comparison fails, decide whether the baseline is stale or the behaviour diverged, and name each intended difference in the test. 
\end{fields}

\card{S5}{Hosting the engine and the link to it}{\TagT}{harnesses with a separate engine process}
\plain{A slow window must not lose text, and an engine that keeps crashing must stop and say why.}
\begin{fields}
\item[Decision.] How is the engine hosted, how do host and engine talk, and what happens when the reader is slow or the engine dies? 
\item[Options.] 
\begin{opts}
\item[Hosting] (\Bel): the interactive terminal program, run in a terminal session the host drives with simulated keystrokes (reuses the existing program; its state is split between the screen and other stores); or a headless child process that reads typed requests on standard input and writes typed events on standard output (a common approach; one place for state; a protocol to define and keep). One process per chat pays the start-up time for every chat.
\item[Message list:] documentation only (no build step; the two sides drift) or one typed table both sides compile against (drift fails the build; a rename touches both sides).
\item[Flow control] when the window reads slower than the model streams: an unbounded buffer (simple; memory grows), dropping (bounded; text is lost), or pausing reading of the model's stream until the reader catches up (nothing is lost; time spent blocked must not count as model silence).
\item[Link:] restart a dead engine with backoff and stop after a crash-loop limit (recovers from a one-off crash; a loop ends with a visible error); a versioned handshake at start (an old engine under a new host gives a plain message; one more step at start); reconnect by numbered events (a returning window catches up without gaps; every durable event needs a number; \Bel).
\end{opts}
\item[Invariant.] (\textup{a}) \inv{with the reader stalled the engine stops consuming the model stream, drops nothing and delivers all text in order when reading resumes}~\Tst{}; (\textup{b}) \inv{an engine that exits a stated number of times in a row (three in our harness) is not restarted again, and the chat shows its last error}~\Tst{}. 
\item[Corner cases.] An engine said to be served over a pipe is really the terminal loop reading fake input, so closing the pipe ends its input reader and the exit fails (S2). A message kind is sent but declared nowhere. A slow reader is counted as model silence; a test that waits for the provider while the pipe is blocked can never pass. The engine exits three times in a row while the chat shows ``Done''. 
\item[Test.] Block the reader for longer than the stall timeout; assert no stall line, no early turn-complete, and intact, ordered text when reading resumes. Start an engine that exits at once; assert one failure line after the third exit and no fourth launch. Start an old engine under a new host and assert a plain message that keeps the queued message. 
\end{fields}

\card{S6}{Cost booking and the budget stop}{\TagT}{harnesses that show cost or cap spending}
\plain{Shown cost matches what was billed, failed attempts included, and a cap really stops the run.}
\begin{fields}
\item[Decision.] How are tokens and money counted when attempts are retried, dropped or switched, and where is a spend cap checked? 
\item[Options.] 
\begin{opts}
\item[Booking:] only the kept attempt (simple; hides attempts that were paid for) or every billed request once (matches the bill; every attempt must be counted).
\item[Unknown prices:] shown as zero (hides spend) or as ``unknown'', with capped runs refusing to start (honest; such a run cannot start).
\item[Token counts:] characters divided by four (free; too low on some wire formats), corrected by the last reply's real usage (true from the second request; the first one after a large attachment is undercounted), or the vendor's counter (exact; one more call).
\item[Budget check:] after a kept reply only (simple; overshoots while tools run) or also after tools and for each account key, since a key can carry its own budget (stops near the cap; more check points).
\item[Helper calls:] booked to the parent chat, so the parent shows the full cost.
\end{opts}
\item[Invariant.] (\textup{a}) \inv{total cost equals the sum of every request actually billed, each counted once}~\Tst{}; (\textup{b}) \inv{no unknown price is shown as zero}~\Tst{}; (\textup{c}) \inv{a capped run stops at or just after its cap and says so}~\Tst{}. 
\item[Corner cases.] A model switch counts usage twice. A switch pays for the cache again. Plans and local models cost zero by design and must not look unpriced. A key carries its own budget, which a check of the run's cap alone misses. Before the first reply returns real usage, a large attachment is undercounted (seen once, \Bel). Background checker and memory calls appear in no task's cost (\Opn). 
\item[Test.] Compare booked tokens with the scripted provider's request log after a failure scenario. Run with an unpriced model id: no zero is shown, and a capped run refuses to start. Run a priced scripted provider with a cap: the run stops at or just after the cap and says so. Return real usage above the estimate; assert the meter uses it from turn two. 
\end{fields}

\subsection{Permission and trust}\label{sec:perm}

An agent that can run commands can delete files, send data away, or change its own rules. The permission layer's difficulties are less about rules than about reading: a command as the shell reads it, paths as a sandbox sees them, a ``read-only'' command that runs a program, a secret by name or by place, a turn that looks watched when nobody is there. Each card pairs a rule with the ordinary work it must not break.

\card{P1}{Modes, rule sources and precedence}{\TagO}{every harness}
\plain{Prohibitions are checked first and win; files written by others never quietly add power.}
\begin{fields}
\item[Decision.] What may the agent do unasked, who may widen that, and in what order are the layers consulted? Rules come from the person, the project (writable by teammates or the sender), the organisation and other tools; \gl{deny rules}, \gl{protected paths}, the mode, hooks, the checker and approval requests can disagree. 
\item[Options.] 
\begin{opts}
\item[Modes:] many, one per ecosystem (precise; confusing) or five or six, each with one name and one plain promise (clear; loses other tools' terms).
\item[Rule sources:] all merged (simple; a repository can grant itself rights), widening only from the person's own files (safe; less convenient), or widening from a project after \gl{folder trust} (a middle path).
\item[Order of layers:] one ordered function (the order of the checks is itself the rule, so a reviewer must read the whole function), a named chain of steps (readable; the order must be kept when it changes), or priorities (flexible; easy to misorder).
\item[Most permissive mode:] drops the sandbox, as the harnesses tested with such a mode do (simple; one click turns any injected instruction into the person's full power) or skips approval requests but keeps the sandbox and hard refusals (safer; tools that start their own sandbox need a retry with approval).
\item[Outside agent programs] (another agent program run inside the harness): the mode is held by that program (no work; a steered agent raises its own mode) or by the harness on every \gl{endpoint} through which the program acts (stronger; \Bel).
\end{opts}
\item[Invariant.] (\textup{a}) \inv{deny rules, protected-path refusals and sandbox-exit refusals are decided before the mode, any hook, the automatic checker or an approval request}~\Opn{} (does not hold in our harness; Section~\ref{sec:cross}); (\textup{b}) \inv{no later layer can turn a refusal or an ask rule into an allow}~\Tst{}; (\textup{c}) \inv{an allow rule or other widening key in a project file takes effect only after the person trusts the folder}~\Tst{}. 
\item[Corner cases.] A hook's ``allow'' overrides a deny rule. A skill's allowed tools pass a deny rule. An old ``always allow'' beats a newer deny. A project file sets a widening key other than allow (extra directories, default mode, sandbox, environment). The model sets a ``disable sandbox'' flag on a call, or writes ``already approved'' in its description. A settings endpoint accepts an agent's credential. An outside agent program's own shell writes a more permissive value into its mode file (\Bel). 
\item[Test.] A matrix of rule kind by mode by hook answer by \emph{flagged} (marked dangerous by the checker): the strongest refusal wins in every cell. Per source file and loader key, plant a loosening value: no change before trust. Write each protected path from every tool in every mode: refused. Call settings and rule endpoints with the person's credential and an agent's: only the first succeeds. 
\item[Cost and limit.] A person who relied on a repository's shared allow rules gets an approval request for each command until they write their own; trusting a folder does activate its allow rules. 
\end{fields}

\card{P2}{Reading a shell command exactly as the shell does}{\TagO}{harnesses that ask before unlisted commands}
\plain{Run a command unasked only if it is read exactly as the shell will read it.}
\begin{fields}
\item[Decision.] When may a command run without an approval request, and how is it read? Approving needs an exact reading, refusing must catch whatever might be dangerous, and one reading cannot do both. 
\item[Options.] 
\begin{opts}
\item[Readers:] one parser for both approving and refusing (simple; cannot be exact and broad at once); a strict reading to approve and a permissive one to refuse, with deny rules matched against the permissive one (safe; more approval requests for odd but harmless commands); or a grammar library (accurate; its own bugs).
\item[What ``simple'' means:] a blacklist of unsafe syntax (fewer approval requests; new syntax is allowed by default) or a small whitelist grammar (new syntax asks by default; more approval requests).
\item[Redirects:] any redirect disqualifies (noisy; trains people to click yes) or only the literal null device and descriptor copies pass (quiet; each allowed form must be read exactly).
\item[Version-control reads:] a status or log command counts as read-only only if (1) no global option redirects its folder or configuration, (2) no change of directory shares the command, (3) the command does not first create the tool's own store files, and (4) neither the folder nor its parents hold the files of a bare repository or a link to a store nobody verified (a repository cannot make a ``read-only'' command run its own program; more approval requests).
\end{opts}
\item[Invariant.] \inv{In every mode that asks, a command runs without an approval request only if the strict reading models every character exactly as every supported shell would; anything else asks}~\Opn{} (does not hold in our harness; Section~\ref{sec:cross}). 
\item[Corner cases.] Substitutions, here-documents fed to a shell, wrappers that execute their argument, command names held in variables and inline interpreter code all run more than they show. A cluster tool's configuration names a program that runs even on a read. Look-alike letters evade deny lists; whether to normalise them is \Opn. 
\item[Test.] Build \emph{near-miss tables}: insert every control and space-like code point at every token boundary of every approved form, plus each shell-dialect oddity, and run them in every mode; none may move from ask to allow. Build a trap folder (a temporary folder laid out with the store files of a bare repository) and assert an approval request with a warning there, while a normal repository's status still runs unasked. 
\item[Cost and limit.] More approval requests for odd but harmless commands; inline code and command names in variables are not read, so in the most permissive mode only the sandbox and hard refusals stand. 
\end{fields}

\card{P3}{What a per-command sandbox can and cannot express}{\TagB}{harnesses with a sandbox}
\plain{A sandbox rule describes paths, not files: block renames and moves too, and write down the gaps.}
\begin{fields}
\item[Decision.] What may a sandboxed command reach, and what can its profile not say? A profile applied to each command matches paths and operations, not file identity or intent. 
\item[Options.] 
\begin{opts}
\item[Profile shape:] default-allow with specific denies (compatible; many holes), default-deny with specific allows (strong; breaks name lookup, databases and container sockets), or a container or virtual machine (strongest; costly, and a separate environment).
\item[Network:] open (compatible; hiding secrets is then the only line against sending them out) or off behind a proxy allow list (strong; installs and fetches need lists). The right default is \Opn.
\item[Tools with their own sandbox:] sandboxes do not nest, so such a tool fails inside; retry it outside behind an approval request (it works; one more question).
\end{opts}
\item[Invariant.] (\textup{a}) \inv{for every denied path the profile also denies creating, renaming and removing that path and its parents}~\Bel{}; (\textup{b}) \inv{this holds for paths that do not exist yet}~\Bel{}; (\textup{c}) \inv{no link target ever appears in an allow rule}~\Bel{}. 
\item[Corner cases.] A login file is renamed into a writable cache and read under its new name. A prepared folder is moved into a place that does not exist yet. A ``hardening'' step that follows links adds the target to an allow rule. A \gl{key agent} is reached through its socket without reading any file. A hard link made earlier reads through. Startup files and hooks written now run later outside the sandbox. 
\item[Test.] For each denied path, from a sandboxed shell: read it, rename it, rename its parent, move a file in from elsewhere, create at the path, and make a symlink, a hard link and a socket; plant links at each fixed place; time command start with the full rule list and fail if it grows. 
\item[Cost and limit.] Tens of milliseconds per command, about 14\,ms more after a long rule list; profile strings over about 1\,KB are rejected, so one rule per entry. 
\end{fields}

\card{P4}{Hiding credentials}{\TagO}{harnesses with a sandbox}
\plain{Hide login keys from every tool; whether a project may read its own secret files is a choice with a cost either way.}
\begin{fields}
\item[Decision.] What is hidden, from which tools, at what cost to ordinary work? Fixed login places (remote-login keys, cloud logins, package-publishing tokens) are secrets by place; environment and key-named files are secrets by name and are also ordinary project files. 
\item[Options.] 
\begin{opts}
\item[What is hidden:] one list hidden everywhere (strong; breaks projects); fixed login places hidden in every mode and never reopened, while name rules stay where they already worked and are tested against previous behaviour (strong where it matters; some secrets by name stay readable); or the person's own list (flexible; little protection).
\item[A project's own secret files:] readable when that project is the working folder (ordinary work keeps running, but an agent steered by injected text can read them) or refused or asked for (safer, and some ordinary work stops).
\item[Tools that read a configuration on every call:] hide only its credential part, or leave it readable and write-protect it (the tool keeps working; that file stays readable).
\item[Environment:] variables with secret-like names, and always the agent's own credential variable, are removed from every command's environment (no leak by name; a tool that needs a token must be given it through an explicit approval).
\item[Engine keys:] passed by pipe or in memory, never through a readable file.
\end{opts}
\item[Invariant.] \inv{Fixed login places are hidden from commands and from every in-process file tool, reads included, in every mode}~\Opn{} (does not hold in our harness; Section~\ref{sec:cross}). 
\item[Corner cases.] In-process read, search, upload or serving tools ignore a sandbox that wraps only child processes. A hidden path arrives under another property name, as a file URL, relative, or nested. Backup forms (``.bak'', ``\textasciitilde'', ``copy'', ``(1)'') are not matched. A forged \gl{linked checkout} points into a hidden place. A hidden global version-control configuration makes every call of that tool fail. A chat in a subfolder cannot read the repository root's environment file. A secret-named variable imported from another tool's settings reaches a command. 
\item[Test.] Walk the tool \gl{registry}: every path-like property, a hidden path in three spellings, every mode; assert ask or refuse. Test 50 or more name variants against profile and file check. Run common tools in a fake home with each entry hidden. Set a canary variable for each secret-name pattern and assert it is absent from a command's environment. Run the \emph{ordinary-work set} (the project's environment file, its tests, a linked checkout, a subfolder chat, a project under a hidden folder) and a \emph{differential test} against previous verdicts. 
\item[Cost and limit.] A same-user process outside the sandbox can read any file the person can. 
\end{fields}

\card{P5}{Who is watching, decided per turn}{\TagB}{harnesses that ask, with schedules or peers}
\plain{A turn nobody watches must refuse what needs a person and must not borrow the person's authority.}
\begin{fields}
\item[Decision.] Is someone present to answer an approval request in this turn, and what may an unwatched turn do? Schedules, webhooks, bots, \gl{peers} and the harness's own nudges start turns with nobody there, and an approval request for nobody waits forever. 
\item[Options.] 
\begin{opts}
\item[Unwatched default:] the most permissive mode (no setup; a scheduled task that reads a poisoned page acts with full shell and network); edits in the project plus read-only commands, refusing anything that would ask (safe; automations need allow rules); everything except flagged commands (fewer refusals; a poisoned page can still send files away); or a sandbox limited to listed sites (strong; needs the list).
\item[Where ``watched'' is marked:] on the chat (simple; one message marks it for good) or on each turn, from the message that started it (exact; every caller must declare it).
\item[\gl{Application text}:] it inherits the last turn's state (simple; it runs with the person's authority) or defaults to nobody, with a short named list that continues a turn: a restart prompt, the continuation of an interrupted turn (safe; every caller declares a value).
\end{opts}
\item[Invariant.] (\textup{a}) \inv{a turn counts as watched only if a message that started it carries the person's mark}~\Bel{}; (\textup{b}) \inv{an unwatched turn refuses at once, saying nobody is there, everything that would ask}~\Bel{}. 
\item[Corner cases.] One message marks the whole chat as watched. Nudges, reports from checks the harness repeats in the background, and limit-reset messages that arrive hours later inherit the person's authority. A scheduled prompt runs a settings command (``add allow rule''), bare, with arguments or under an alias. An agent schedules a child run with a stronger mode. A permissive chat keeps running after the person leaves. 
\item[Test.] (1) A person's turn, an application text turn, a peer turn: an approval request, a refusal, a refusal. (2) Every caller of application text, after a watched and an unwatched turn: its declared value. (3) Every command under every name: fail any without a ``changes settings'' declaration. (4) Every ask path from an unwatched turn ends at once. 
\item[Cost and limit.] Automations must be redesigned to decide without asking; an overnight job in a permissive mode that hits a limit is resumed in the capped mode and refuses commands unless a rule allows them. 
\end{fields}

\card{P6}{Extensions that grant themselves rights}{\TagT}{harnesses with skills, hooks or tool servers}
\plain{Nothing in a folder may run or grant rights until the person trusts that folder.}
\begin{fields}
\item[Decision.] What may skills, commands, hooks, tool servers and settings imported from other tools grant? Each is a file someone else can write that names a program or allows a tool. 
\item[Options.] 
\begin{opts}
\item[Whose grants count:] all (simple; any repository grants itself rights); only the person's folders, built-ins and installed plugins, in watched turns, with repository hooks and servers active only after folder trust and bound to a hash or the exact approved command, and other tools' allow rules shown greyed until the person confirms an import (safe; shared project skills ask per tool); or other tools ignored entirely (simple; their deny rules are lost too).
\end{opts}
\item[Invariant.] (\textup{a}) \inv{nothing a folder's files declare runs or grants before the person trusts that folder}~\Tst{}; (\textup{b}) \inv{extension grants come only from resolved folders owned by the person or the harness}~\Tst{}; (\textup{c}) \inv{they count only in watched turns (as P5)}~\Bel{}; (\textup{d}) \inv{allow rules from another tool's files never run a command unasked}~\Tst{}. 
\item[Corner cases.] A prefix test lets ``home-evil'' pass as ``home'', and a comparison without a separator does the same. A symlink leads out of a trusted folder. A hash covers the hooks configuration but not the scripts it names (\Opn). Another repository's linked checkout with identical hooks inherits trust. A tool server's command changes and keeps its approval. A hook whose answer comes from a model returns a malformed answer (\Opn). 
\item[Test.] For skill grants: look-alike, symlinked and linked-checkout folders and unwatched turns. For hooks: change a hook, then only its script, then revoke trust. For servers: plant a changed command. For imports: put allow, deny and environment entries in a copy of another tool's settings and assert only deny and ask apply. 
\item[Cost and limit.] Scripts named by a trusted hook are unprotected between trust and edit; imported workflows may need re-entry. 
\end{fields}

\card{P7}{Approval requests}{\TagT}{every harness that asks the person}
\plain{An approval request shows the harness's own warning first and never answers itself.}
\begin{fields}
\item[Decision.] What must an approval request show, where may it be approved, how is an answer matched, and what happens to approval requests left open? The model writes the description, \gl{surfaces} differ, and several requests can be open. 
\item[Options.] 
\begin{opts}
\item[Content:] the model's description only (simple; a dangerous command can carry a calm reason) or the checker's flag first, in its own labelled row (honest; one more row).
\item[Surfaces:] one tap everywhere (fast; too easy) or one tap only for unflagged requests, with notifications on a locked device able only to deny (safer; slower).
\item[Matching:] one ``current request'' slot (simple; an answer can hit a request the person never saw) or an id per request, refused if unknown or already answered (exact; common in four harnesses tested, \Bel).
\item[Waiting:] a timeout, then deny (no stall; answers for the person); no timeout (never answers for the person; the turn stalls); or no timeout plus one reminder after a delay (never answers for the person; the turn can still stall, but the person is told once).
\end{opts}
\item[Invariant.] (\textup{a}) \inv{an approval request shows the harness's own reason before the agent's words}~\Tst{}; (\textup{b}) \inv{while the harness runs, it is never answered automatically}~\Tst{}; (\textup{c}) \inv{a stop withdraws every open approval request, so that a later answer is refused and the model is told nobody refused}~\Tst{}. 
\item[Corner cases.] A calm model-written reason sits on a dangerous command. Hidden or direction-changing characters disguise the command; show them as codes and scroll to the end. A flagged request is approved from a notification. An answer lands on a request the person never saw. A request open at restart is answered ``cancel'' by the dying process and read as a refusal; whether requests should survive a restart is \Opn. Harmless approval requests push people toward the permissive mode (seen once, \Bel). 
\item[Test.] A flagged ask carries and shows the reason first. Each surface with flagged and unflagged requests. Stop with a request open, then Allow: refused. Quit with a request open and reopen: expired, and the model told nobody refused. Reminders on a fake clock. With ids: answer the second of two requests by id, then replay it. 
\item[Cost and limit.] An extra row and slower phone approval; four of five harnesses tested wait without a limit. 
\end{fields}

\card{P8}{Standing grants and the text of a refusal}{\TagT}{every harness that asks the person}
\plain{``Always allow'' lives where the model cannot write; a refusal says who refused.}
\begin{fields}
\item[Decision.] What does ``always allow'' save, how wide, and where; and what does the model read when an action is refused? A grant the model can write lets it grant itself more; a bare ``denied'' invites retries and workarounds. 
\item[Options.] 
\begin{opts}
\item[Width:] the exact command (safest; repeat approval requests), a prefix with a visible ``covers'' line (wider; fewer requests), or a whole tool (widest; covers anything the tool can do).
\item[Storage:] a file only the person writes (safe; one more file), a project file (shared; others may write it), or session memory (no file; lost at restart).
\item[Leaving the sandbox:] one yes per call; a standing grant only for the byte-identical command (a changed wrapper or host asks again; more approval requests).
\item[Refusal text:] a bare ``denied'' (short; invites retries and workarounds) or one naming the refuser (person, rule, mode, nobody present, safety check), saying the action was not performed and, for hard rules, forbidding workarounds (clear; longer).
\item[\emph{Breaker}:] end the turn after several refusals (stops loops; cuts off alternatives; the right count is \Opn).
\end{opts}
\item[Invariant.] (\textup{a}) \inv{a saved grant is written only to a file the model cannot write in any mode, and shows what it covers before it is saved}~\Tst{}; (\textup{b}) \inv{a grant to leave the sandbox matches only the byte-identical command}~\Tst{}; (\textup{c}) \inv{every refusal text names its source and says the action was not performed}~\Tst{}. 
\item[Corner cases.] On a phone, ``always allow'' should count as once, because the covering text cannot be read there. A settings command in an unwatched turn plants an allow rule. A sandboxed command can write the home settings file. Text in a call claims prior approval. A wrapper changes the bytes of an approved command. A refusal that does not say who refused, or that nobody was there, is routed around by the model. 
\item[Test.] Write the grant file from file tools and a sandboxed shell in every mode: all fail. Save an exact grant, change one byte (space, host, wrapper): a new approval request. Every refusal path names its source and says ``not performed''. A model asking for a refused tool past the breaker's count: the turn ends with a visible row. 
\item[Cost and limit.] Whether models retry a refused action in another spelling is \Opn; the text discourages it but cannot prevent it. 
\end{fields}

\card{P9}{Other agents' text and outside content}{\TagO}{harnesses with peers or web content}
\plain{Outside text may contain instructions; it is \gl{fenced} as data and never counts as the person's consent, though it can still sway the model.}
\begin{fields}
\item[Decision.] How is text from other agents, web pages or tool servers admitted, and whose authority does it carry? Its author may be steering it, and a peer's message in a watched turn would ride on the person's presence. 
\item[Options.] 
\begin{opts}
\item[Admitting outside text:] pass it raw (simple; anyone who writes it can steer the agent); fence it as data with one marker (cheap; the model may still be influenced); or strip invisible characters and bound its length, or refuse it (removes hidden instructions; refuses some legitimate text).
\item[Loops between agents:] refuse a body repeated within about ten minutes and stop chains at a hop limit (stops loops; refuses legitimate repeats).
\item[Accepted trade-off:] other agents' text that arrives during a watched turn is part of that turn, because a task the person started must be able to read its helpers' replies; so a watched turn can carry a request a peer injected.
\end{opts}
\item[Invariant.] (\textup{a}) \inv{text from other agents and from outside content is fenced as data}~\Opn{} (does not hold in our harness; Section~\ref{sec:cross}); (\textup{b}) \inv{it never counts as consent for the checker}~\Tst{}; (\textup{c}) \inv{it never changes whether a turn is watched (as P5)}~\Bel{}. 
\item[Corner cases.] Two agents trigger each other in a loop. Hidden tag characters carry instructions. The same body is sent again and again. A held note from a helper rides in with the person's next message. Two fence styles with two jobs (one for a summariser's input, one for tool results) are mistaken for one. A rule is adopted that text in reminder tags is authoritative, while recalled memory sits in those same tags. 
\item[Test.] Hidden characters, repeated bodies and a three-hop chain: stripped, refused, stopped. Instructions in a web page and a tool result fed to an obedient scripted model: fenced, never counted as consent. A peer message mid-turn, watched and unwatched: the watched state does not change. 
\item[Cost and limit.] The fence does not stop a model from being influenced; whether fences change behaviour under injection is \Opn. 
\end{fields}

\subsection{What the person sees}\label{sec:view}

The person judges the harness by the screen. Two rules carry most of the weight: each event is announced once, by one component, in words meant for a person; and whatever a helper agent does reaches the parent chat as plainly as the parent's own work.

\card{V1}{One notice per event, failures with their own name}{\TagT}{every harness}
\plain{Each event is announced once, in plain words, with true numbers, and every ending has a reason.}
\begin{fields}
\item[Decision.] Who words what the person reads about an event, and what must every final state show? Several surfaces want to announce each event, and model-facing text leaks. 
\item[Options.] 
\begin{opts}
\item[Wording:] each layer words the event (no coordination; one event, three announcements) or one function per event kind with one title per failure code (consistent; every code needs a title).
\item[Numbers:] computed by the window (simple; wrong counts) or sent as fields from the record (true; more fields to send).
\item[Where notices go:] to the log only (quiet; the person never learns that old results left the model's view) or through one helper with three classes: a normal row, a faint row when the harness nudged the model, and log only when nothing changes for the person (every notice is placed on purpose; each code needs a class).
\item[Chat status:] derived from timestamps by many writers (simple; wrong in corners) or computed by one function (testable; every state needs a case).
\item[Audience:] text for the person and text for the model are kept apart for each event kind.
\end{opts}
\item[Invariant.] (\textup{a}) \inv{each event produces at most one notice, with one title from a shared function}~\Tst{}; (\textup{b}) \inv{every number in it is a field from the record that drove the behaviour}~\Tst{}; (\textup{c}) \inv{every final state a chat can reach has a reason a person can read}~\Tst{}; (\textup{d}) \inv{a count shown counts what the person cares about, requests sent or time spent, not one layer's attempts}~\Opn{} (18 requests under overload were shown as at most ``try 2 of 10'', the counter starting again in each waiting round, and logged as ``after 3 attempts''; F1, Section~\ref{sec:cross}). 
\item[Corner cases.] One stop is announced three times. One failure gets two titles. ``Ran \ldots, failed'' is drawn for a command that never started. A hook that succeeds after printing an error is drawn as ``success: connection refused''. ``Done'' shows while messages are queued and no error is given. ``Nothing running'' shows while a chat waits for approval. A refusal is announced as an empty response, and a command as timed out although it ran to completion (our harness, Section~\ref{sec:cross}). 
\item[Test.] Every notice code has a record and a person-text key, and a source search finds none emitted outside the helper. Count rows for scripted events. Every failure code has a title; the row verb is right for denied, stopped, failed and ran (the denied case is not yet tested). Every final state has text. Five polls with a lasting cause give one notice. 
\end{fields}

\card{V2}{Helper agents}{\TagT}{harnesses with helpers}
\plain{A helper works for its chat: its result, failure, cost and stop all show up there.}
\begin{fields}
\item[Decision.] What does a helper receive, how does its state reach the parent, and what happens when the parent stops? A helper is a second loop whose failures, cost and processes belong to the parent. 
\item[Options.] 
\begin{opts}
\item[Context:] forked with the parent's history (the helper knows the background; it also inherits the parent's tools, which must still work inside it: a helper that wrapped the parent's already-wrapped tools reported a loaded tool as ``not loaded'', observed in another harness and not reproduced, \Bel) or a fresh brief (focused; a long brief can be cut by the rule that shortens very long messages, seen once, \Bel).
\item[Model:] inherited unless the person asked for another (works on every family) or a fixed id (simple; fails, or silently runs elsewhere, on other families).
\item[Display:] events tagged by agent id on one stream, a roster of states (queued, working, suspended, completed, failed), and counts derived from the states (consistent; every view must filter by id).
\end{opts}
\item[Invariant.] (\textup{a}) \inv{a helper's status shows its recorded outcome, with counts derived from states}~\Tst{}; (\textup{b}) \inv{its cost is booked to the parent}~\Tst{}; (\textup{c}) \inv{stopping the parent stops it and says who stopped it}~\Tst{}. 
\item[Corner cases.] A helper shows ``Task failed'' and ``0 agents completed'' while the real reason sits in a saved report. A helper waiting for approval is shown as failed. A helper is pinned to a model another family lacks. A command a helper started early survives the stop. Helpers ask for approval that nobody sees (\Bel: the policy for it is from comparison). 
\item[Test.] Script a helper that fails, one that waits for approval and one that finishes; assert the status text and counts. Start two helpers, stop the parent, and assert no live child process and result texts naming who stopped them. Sum the usage records and compare with the cost shown for the parent. 
\end{fields}

\subsection{How ten other harnesses decide}\label{sec:compare}

Our harness's own choices are in the cards. For comparison ten other harnesses were manually tested and their behaviour observed, among them command-line agents and desktop applications; open-source harnesses were also studied as reference. Versions and dates were not recorded, so every statement drawn from them is \Bel{} and is recollection of the testing, not a checkable record. ``A common approach'' in the cards means one seen in several of them; exact counts are given only where known (for approval requests five showed the decision, and four of those wait without a limit; from the author's notes). Some options in the cards were adopted from these observations and are marked as from comparison. Separately, two other harnesses were also run against the checklist during this work, and at least one of them is among the ten tested; their results are not reported (Section~\ref{sec:cross}), and a few observations from those runs appear in the cards, marked as seen in other harnesses. They converge where a constraint is clear: read-only calls start during the reply and rerun after a retry; none of those with a most permissive mode keeps the sandbox in it; none checks at runtime a claim the model makes with no call behind it. They diverge where it is not: silence is detected after about 30\,s in some, by a colour change at 10\,s or a line at 60\,s in others; a summary is triggered at half the context limit in one and 85\% in another; a stream with no stop reason is retried by some and kept and continued by others. The same pattern appears in a source-code taxonomy of scaffolds~\cite{rombaut}. 

\section{Failures by cause, and four worked cases}\label{sec:catalogue}

The corner cases above repeat a few mistakes. Sorting them by cause lets a builder who finds one look for its siblings. We sort them into six cause classes: \emph{two readers} of one thing that disagree; an \emph{unlisted path} no rule reaches; \emph{untrue text} told to the model or person; \emph{loose state} kept in several places or not reset; \emph{borrowed authority} not granted by the person; and a \emph{blind test} that cannot fail or is never run. Table~\ref{tab:causes} maps the classes to the cards whose corner cases show them, with the cheapest general detection. The classes are a reading aid, not a proven taxonomy (\Bel); coded failure modes for multi-agent systems~\cite{cemri} and failure attribution~\cite{zhang} are the models.

\begin{table}[tb]
\centering\footnotesize
\renewcommand{\arraystretch}{1.25}
\caption{Six cause classes, with the cards where each appears and the cheapest general detection.}\label{tab:causes}
\begin{tabularx}{\linewidth}{@{}>{\raggedright\arraybackslash}p{0.17\linewidth}>{\raggedright\arraybackslash}X>{\raggedright\arraybackslash}p{0.17\linewidth}>{\raggedright\arraybackslash}p{0.27\linewidth}@{}}
\toprule
\textbf{Cause class} & \textbf{What it is} & \textbf{Cards} & \textbf{Cheapest detection}\\
\midrule
Two readers & Two components read one thing differently: checker and shell, prompt and tool list, timer and record & T1, T4, T5, F3, C1, C3, S5, P2, P3, P6, V1 & Feed both readers the same generated inputs and compare\\
Unlisted path & An exit, caller, tool, key or output mode that no list enumerates, so no rule reaches it & T1, T3, T5, F2, C3, S2, P4 & Walk the registry or the syntax tree; one test per entry\\
Untrue text & Something the model or person is told that is false, often a guess shown as fact & T3, F2, F3, F4, F5, C1, S2, S6, P7, V2 & Assert exact texts per cause; scan for guessed values\\
Loose state & A counter, clock, queue or id kept in several places, matched by position, or not reset & T2, F1, C2, S1, S3, S6, V1 & Fake clock; count requests and rows; kill and restart\\
Borrowed authority & A turn, file, link or agent acting with rights it did not get from the person & P1, P3, P5, P6, P8, P9 & Plant the widening input; assert nothing changes\\
Blind test & A test that cannot fail, is not run, or runs a mode no real use reaches & F4, S4, P4 & Plant changes; count tests discovered against tests run\\
\bottomrule
\end{tabularx}
\end{table}

\subsection{Four worked cases}\label{sec:cases}

\paragraph{Case 1: the checker and the shell read one command differently (P2).} To spare people harmless approval requests, a harness lets a short list of forms run without asking, read by a reader that was written to find a command's name. Where that reader and the shell disagree about what belongs to a word, a form the reader takes for a harmless one is, to the shell, another command. The reader was not built to prove what the shell will do, and was then trusted to approve. A review pass found the class. \emph{Rule:} approval needs a strict reading that models every character as every supported shell does, and every loosening ships with a near-miss table generated around the new form. 

\paragraph{Case 2: a security change that breaks ordinary work (P4).} A change hid credential files from commands and file tools, by name as well as by place, and passed a security review pass. Asked to test ordinary work, the same kind of pass found that a chat in a package subfolder could not read the repository root's environment file, a linked checkout was refused for a tracked registry settings file, and projects under a hidden or synchronised folder lost their own environment file. Secrets by place and by name had been treated as one, and the review measured only what was blocked. \emph{Rule:} every security change runs a fixed set of ordinary tasks and a differential test of previous verdicts, as one injection benchmark measures utility beside attack~\cite{agentdojo}. 

\paragraph{Case 3: a test mode nobody runs (S4, C2).} Our engine runs directly and served through a pipe, with recorded baselines for both. The default test run covered only the direct mode, so baselines and test files of the served mode failed unnoticed while every check that ran was green, and a prompt-size test ran in a mode no real chat uses. \emph{Rule:} every test that exists is run by the automated checks, a failing test blocks merging, and each baseline is tied to a mode real use reaches. 

\paragraph{Case 4: a prompt fix that does not remove a habit (F4, T1).} A small model wrote its closing, then, asked again after a finishing tool, wrote it again in other words. In ten real runs per option, a prompt sentence against repeating did not lower the rate, nor did also rewording the tool's result; a display rule removed it (Appendix~\ref{app:evidence}, F4). A prompt sentence makes a habit more or less likely; it does not remove it. \emph{Rule:} measure any model-facing wording change on real runs before and after, and where the harness can observe the event, prefer a runtime or display rule. 

\section{Testing a harness}\label{sec:testing}

The fault scripts, recordings and tests described here were built around Crewroom's engine. Most card tests need the same tools: a fake provider that fails on demand, a recording of what the harness did, and proof that the recording notices a change. The method runs in order: state the invariant as a testable sentence; build a scripted provider, fault relay and fake clock; record behaviour for every scenario and transport; plant changes until each is caught; confirm three runs are byte-identical; only then change code under the recording; then add probes, the ordinary-work set, a check on screen and one real run per model family (the recorded real runs of this paper cover one family only). Each kind of test also has a blind side: a table test misses cases not listed and decisions made outside the function; a scripted provider misses real timings, wording and model habits; a behaviour recording misses behaviour that was already wrong and branches no scenario reaches; a registry walk misses dangerous behaviour of a declared entry; probes miss the ordinary work a change breaks; a check on screen with a scripted provider misses the real service; and one real run per family misses faults that cannot be forced.

\paragraph{Scripted providers and fault injection.} A scripted provider per wire format replays chosen reply shapes; a fake clock replaces time, timers and randomness, so minutes of retries run in milliseconds. For \emph{fault injection}, a relay in front of a real provider injects one named fault per request (slow first byte, stall, drop, repeated overload, rate limit with a reset) without logging headers or bodies. A provider that only returns complete replies reaches no failure branch: every failure decision site must be hit by a scripted row, and every failure scenario should run on every wire format, not only the busiest. Ours falls short: its 150 fault rows give 187 row-and-wire runs (114, 47, 22 and 4 on its four wire paths), so most rows run on one. 

\paragraph{Record behaviour before restructuring, and prove the recording is sensitive.} Turn functions grow past a thousand lines, with one fact (``how long have we been failing'') held in loose variables reset by hand in several places. Before moving code, record everything observable per scenario and transport: request bodies, events, log lines, frames, output streams, exit codes. Each move must leave the recording byte-identical or be reverted. To prove the recording can notice a change, plant one-line mistakes in a throwaway copy (a retry maximum off by one, a reminder word changed, a frame skipped); a \emph{planted change} that nothing catches means a missing scenario. Building such a recording, running it three times and proving it sensitive this way is tested; moving code under it is not yet exercised, so the move-and-compare half is \Bel. 

\paragraph{Determinism traps.} Three runs of the same code must record identically. Parallel results recorded in completion order differ from run to run; record them by call id. A date normaliser and the engine in different time zones disagree at midnight. Absolute paths, drifting counters and token estimates change between machines and runs. Tool servers race the first request. A test that counts same-millisecond arrivals as ``after'' is flaky; count by sequence number. Run each scenario three to five times, also across midnight. 

\paragraph{Three levels of proof.} A scripted provider proves the logic on each code path for every family that uses it. A check on screen in the installed application, with a failing scripted provider, proves what the person sees. One small real run per family proves the real service and the model's habits. Unit tests passed while the installed application was wrong, for these causes: no test drove the row on screen, a hosting setting was absent from the tests, a fake clock hid a real wait, the build used another runtime, a test asserted old text, and a sentence meant for the model was drawn to the person. 

\paragraph{Adversarial review that also runs ordinary work.} A review pass with its own probe corpus (hundreds of strings in every mode) reports each must-fix as a reproducible probe and re-runs the probes after the fix. In our harness no person outside the project reviewed it. Every added mechanism gets its own attack probe, because a hardening step can open a hole (P3); every fix names its class of bug and tests the whole class; the same pass runs the ordinary-work set (case~2); and the result lists what was not checked. 

\paragraph{Registry walks and differential tests.} Tests that enumerate every registered tool, command, settings key, turn exit, notice code and final state, failing any entry without a declaration or its ask or refusal, keep up with the code: a new tool whose path parameter is not called ``path'' is caught. A recorded table of verdicts (for a permission layer, thousands of cells) is re-run after every change; unchanged rows keep their verdicts, changed rows are listed and approved, and each updated fixture is explained by the decision it follows. 

\paragraph{Re-checking when the model changes.} This is practice, not a tested claim. Parts of a harness written to make up for a weaker model, such as nudges after an empty reply, word lists that classify a reply or an error, and forced retries and re-asks, can become the limit with a stronger model, acting on replies that needed no help. When the model changes, list these parts, measure each on real runs with the new model as F4 asks for prompt fixes, and remove those that no longer earn their cost.

\subsection{Runs behind a fault relay}\label{sec:realruns}

To reach the real service without waiting for an outage, a turn of our harness was run as the host runs it, against a small current model at a hosted model provider, with the engine's provider address pointed at a local relay. The relay forwards each request and can inject one named fault: delay the first byte, stall after a number of stream events, cut the connection, answer an overload status a set number of times or always, answer a rate limit with a retry-after, or return an empty reply. Overload, rate limits and empty replies are produced by the relay; the provider is never asked to fail, and each real request carries a one-line prompt. A runner reads the transcript and engine status as the chat does and stamps each new row; it does not see a row updated in place. No clock was shortened. A first session ran seven scenarios once each, six with a fault and one without; each reached the model, and all seven ended as the cards say, with two notes (the raw transport text after a retry row, and the waiting line during overload seen only in the engine log). A second session ran nine scenarios three times each, one at a time (Table~\ref{tab:realruns}).

\begin{table}[tb]
\centering\footnotesize
\renewcommand{\arraystretch}{1.15}
\caption{Runs of our harness behind a local fault relay, second session: nine scenarios, three runs each, one small model, one wire format. In the two rows marked ``relay only'' the relay answered every request, so no model was involved.}\label{tab:realruns}
\begin{tabularx}{\linewidth}{@{}>{\raggedright\arraybackslash}p{0.2\linewidth}>{\raggedright\arraybackslash}X>{\raggedright\arraybackslash}p{0.15\linewidth}>{\raggedright\arraybackslash}p{0.14\linewidth}@{}}
\toprule
\textbf{Fault injected} & \textbf{What our harness did} & \textbf{Wall time, s} & \textbf{As card?}\\
\midrule
None & One request, reply & 1.7, 1.3, 1.5 & yes\\
Cut after 2 events & Retry row ``connection dropped, trying again in 1\,s (1 of 10)'', second request, reply & 2.3, 3.9, 5.1 & yes (F1)\\
First byte delayed 30\,s & Waiting line at 20\,s naming the reconnect time, cleared when output came; one request & 31.1, 31.3, 31.3 & yes (F2), but its countdown rose\\
Stall of 60\,s after 1 event & Waiting line at 21\,s, cleared at 61\,s; one request (60\,s is under the 300\,s silence limit) & 61.5, 61.3, 61.5 & yes (F2)\\
Second request stalls 60\,s after a tool call & Tool row, waiting line at about 22\,s, cleared; reply & 61.7, 62.8, 62.4 & yes (F2)\\
Overload 3 times & Retry row; 4 requests; a wait of 17 to 18\,s after the third; reply & 19.2, 19.4, 20.2 & yes (F1)\\
Overload on every request (relay only) & Five waiting rounds, 18 requests, final line ``busy right now \ldots\ saved \ldots\ send `continue' \ldots'' & 258.2, 262.1, 259.9 & line yes, time \textbf{no} (F2)\\
Rate limit, retry-after 60\,s & Retry row ``trying again in 60\,s''; 2 requests; reply & 61.7, 64.5, 63.6 & yes (F3)\\
Empty reply on every request (relay only) & One request, no re-ask, no row; failed turn with a plain line naming the next step & 0.3, 0.3, 0.3 & yes (F4)\\
\bottomrule
\end{tabularx}
\end{table}

Of the 27 runs, 21 had at least one request answered by the model, and those 21 ended as the cards say; in nine of them a line differed from the card (three with a countdown that rose, three with a rate limit worded as overload, three with raw transport text after a retry row, Table~\ref{tab:realruns} and the notes below). In the six runs of the two rows marked ``relay only'' the relay answered every request, so no model was involved; both findings below come from them. No run was stopped by the spend guard; run-to-run differences were timing only.

\emph{The give-up time.} Under overload on every request, all three runs gave up after the fifth waiting round, about 260\,s in, while the ``still busy'' lines in the engine log named a stop time ten minutes after the start, about six minutes later. The runner logged only new rows and saw these lines only in the log. A replay answered that question: with a scripted upstream in place of the relay, it re-read the whole transcript every 0.2\,s with the chat's own parser, kept only the latest retry row as the chat does, and rendered it with the chat's own strings and countdown. The retry row is one row updated in place, and its text with the stop time stood for about 230 of the 244\,s; the three runs ended 5.0, 5.2 and 5.1 minutes before the stop time shown. The replay uses the chat's code, not a capture of the screen. In the one-shot runs of Section~\ref{sec:cross} the generic route named a stop time 60 minutes after the start and ended after 286\,s. So the stop time is shown and is false, because five rounds and a ten-minute deadline both bound one episode and the row states only the deadline; card F2's clause (b) is Open.

\emph{The empty reply.} With an empty reply on every request our harness did not re-ask: one request, no row, and a failed turn after 0.3\,s whose error has the kind ``invalid request''. This is a coded rule for this family: an empty first reply that does not follow tool results ends the turn at once with a plain line, and ``invalid request'' is that rule's own label, not a misreading of the relay's reply. In our code ``nudge once'' is the rule for an empty reply after tool results. The replay confirmed both on three empty shapes, the relay's and two built from the provider's documented stream format: 9 of 9 runs failed at once on an empty first reply, and after a tool result 3 of 3 runs nudged once with a row and then got the answer. Other families resend after 5\,s and then 10\,s with counted rows. Card F4 states the rule.

Smaller notes: a rate limit was worded ``is busy'', as for overload; the raw transport error followed the dropped-stream retry row in the runner output, which may or may not reach the window; and in the first-byte runs the waiting line's countdown to the reconnect rose from 4\,min 43\,s at 20\,s to 5\,min 0\,s at 31\,s, so it did not show the true remaining time F2's test asks for. Not run: a refused connection (excluded by the safety rules for these runs), a reply cut at the output limit and a malformed event (the relay has no such fault), a retry-after above a minute (only 60\,s was tried, and retried), and scenarios that need the window. No real outage or rate limit was caused.

\subsection{What is needed to repeat the tests}\label{sec:repeat}

\paragraph{Period and size.} The harness's test base holds 1,126 test files; a text search finds 9,539 places where a test opens, which undercounts table-driven tests and is not a run count. No whole-suite pass count is recorded. The tags are a snapshot of the notes within this period. The evidence was taken at four commits, not all on one branch, in a span of two days; the tags and counts are a snapshot of that span and nothing was re-run after the text was final: the code review, the two relay sessions, and the self-assessment runs of Section~\ref{sec:cross} with the lock count below.

\paragraph{The behaviour lock.} One command runs four groups, prints rows, matches, differences and time per group, and fails if any recorded output differs (Table~\ref{tab:lock}). One run took 242 to 261\,s in the recorded runs, and 250\,s at the commit of Section~\ref{sec:cross}, where it was green. Repeated runs were identical: three runs of the fault group, five byte-identical runs of the pipe traces, three green runs of the whole command, each parallel scenario ten times, and one pipe scenario set 20 times of 20. Planted changes were caught: in the fault group, changing one retry rule's value from four to three, and a changed variable name; in the pipe traces, three planted engine changes turned 35 of 39, 8 and 16 situations red (the record does not say why 39 rather than the group's 40); in the whole command, swapping the order of two parallel results turned 20 rows red. A count at the commit of Section~\ref{sec:cross} put a probe before each of the 62 exits of the turn function (31 breaks, 21 continues, 4 returns, 6 throws) and ran the whole lock: it reached all 62, 61 of them through the fault rows; the parity recording alone reached 16. It also reached all eight verdict kinds of the deciding function of F1 and all eight action kinds of the function that judges a reply's end.

\begin{table}[tb]
\centering\footnotesize
\renewcommand{\arraystretch}{1.15}
\caption{The four groups of the behaviour lock. Units differ: the fault group counts rows and row-and-wire runs, the others scenarios or situations.}\label{tab:lock}
\begin{tabularx}{\linewidth}{@{}>{\raggedright\arraybackslash}p{0.19\linewidth}>{\raggedright\arraybackslash}p{0.15\linewidth}>{\raggedright\arraybackslash}X>{\raggedright\arraybackslash}X@{}}
\toprule
\textbf{Group} & \textbf{Size} & \textbf{Records} & \textbf{Made comparable by}\\
\midrule
Fault rows, in the same process & 150 rows; 187 row-and-wire runs, plus one test that every replaced global is restored & Every event, every request (method, path, version headers, body), the saved record, the final message list, any error & A virtual clock for dates, timers and intervals; random numbers fixed at one half; fresh temporary home and folder; one time zone; the network call replaced by the scripted provider\\
Parity & 65 scenarios & Requests sent, saved record, tool results, terminal text & A scripted provider replaying each scenario; a normaliser for paths, ids, times, durations, system strings, commit ids, image bytes and process-based ids\\
Parity over the pipe & 41 of those scenarios (those a pipe can drive) & The same, plus the frames exchanged, in order & The same normaliser; its own baselines\\
Pipe traces & 40 situations & Every frame sent and received, saved records, exit code and signal & Timed actions wait for a named frame, never a sleep; parallel frames sorted into call order\\
\bottomrule
\end{tabularx}
\end{table}

\paragraph{How a fault row is scripted.} A row has an id, a one-line description, the wire formats it runs on (a message format, a chat-completions format, an item-based format, and a shared path for compatible third-party hosts), optional settings, and a script with one answer per request in order. An answer is a status with headers and body; streamed events, possibly malformed, closed, stalled or reset after $n$ chunks; headers with no body; a connection error; a request never answered; or a Stop. Every answer can carry a latency (500\,ms by default), and helpers build well-formed replies per wire format, so a row states the reply's shape, not its bytes. The pipe traces use a local server that replays scripted replies, holding a reply until a named gate opens, so that timing never depends on a sleep.

\section{A checklist for builders}\label{sec:checklist}

Answer each item, or each lettered part, yes only with a test to show. A star marks the ten cards to do first; terms are in the glossary. Items 21 to 24 (P2 to P5) apply only to the harnesses their card's heading names. Nine lettered parts have no tagged clause behind them in their card: T1(d) and (e), F2(a), F4(c), C2(b), P4(b) and (c), P5(b) and (d); Section~\ref{sec:cross} scores them like the rest.

{\footnotesize
\begin{multicols}{2}\raggedright
\begin{enumerate}[leftmargin=2.2em,itemsep=0pt,topsep=2pt,parsep=0pt,label=\textbf{\arabic*.}]
\item[] \textbf{The turn}
\item $\star$ (a) Every exit of the turn is listed and has a scripted test. (b) Every end except a normal finish shows its reason. (c) Exactly one closing message is stored and drawn. (d) A stream without a stop reason, or a cut reply, is never shown or stored as complete; it is shown as incomplete, withdrawn, or continued with the partial text kept and marked. (e) Every continue path counts against the step limit. [\cd{T1}]
\item $\star$ (a) Results are matched to calls by id. (b) They are stored in call order. (c) After every turn no item is open and results equal calls. [\cd{T2}]
\item (a) Only read-only calls start before the reply ends. (b) An abandoned attempt is withdrawn from every output mode, or kept, marked and continued; it never leaves an orphan result or process. [\cd{T3}]
\item (a) Stored and drawn order equals arrival order. (b) No reasoning appears in the visible reply. [\cd{T4}]
\item $\star$ (a) Every tool result is non-empty and bounded, with the full output retrievable. (b) Every command past its timeout is killed with a result that says so, or tracked until a finish notice. [\cd{T5}]
\item[] \textbf{When things fail}
\item $\star$ (a) One stated bound covers a failure episode's time and attempts. (b) No layer retries what another layer retried. [\cd{F1}]
\item $\star$ (a) On every route a waiting line appears after a stated silence counted from sending. (b) A silent provider ends at a deadline with a plain line that names the next step. (c) The stop time shown is the real one, whichever bound ends the episode. [\cd{F2}]
\item (a) Usage limits, rate limits and overload are told apart using recorded real error bodies. (b) No reset time is shown unless the provider stated it. (c) Every reply by a model the person did not pick is labelled. [\cd{F3}]
\item (a) Every weak-model guard has a bound. (b) Each re-ask, nudge, stop and cut shows its true count. (c) Prompt-wording fixes are measured on real runs before and after. [\cd{F4}]
\item (a) A refused or blocked reply is never shown as an answer. (b) Its unstarted calls never run. [\cd{F5}]
\item[] \textbf{What the model is told}
\item (a) The built prompt for each family has no claim false for the session. (b) No two sections contradict. (c) Every tool named in the prompt is in the same request's tool list. [\cd{C1}]
\item (a) The request prefix is byte-identical from turn to turn. (b) Changing values are appended, not placed in the prefix. [\cd{C2}]
\item (a) Rules come back from their files after a summary. (b) The person's typed words survive verbatim up to a cap. (c) Measured facts survive three summaries. [\cd{C3}]
\item[] \textbf{State}
\item $\star$ (a) A call with side effects is written to the log before its tool starts, and a result before it is shown. (b) After a kill the log can be read and every call has one result, or the chat waits for the person to resolve the unknown outcome. (c) A stopped turn is never continued by a restart (\Bel). [\cd{S1}]
\item $\star$ (a) After Stop no new tool or request starts. (b) Open approval requests are withdrawn. (c) The model is told the truth. (d) The process exits cleanly with hooks run. [\cd{S2}]
\item Each message sent mid-turn is delivered once, shown once, and survives a restart. [\cd{S3}]
\item Every scenario runs in every transport in the automated checks, and every difference is listed. [\cd{S4}]
\item (a) A stalled reader loses no text. (b) An engine that exits three times in a row is not restarted again. [\cd{S5}]
\item (a) Each billed request is counted once. (b) No unknown price shows as zero. (c) A spend cap stops the run. [\cd{S6}]
\item[] \textbf{Permission and trust}
\item $\star$ (a) Deny, protected-path and sandbox-exit refusals come before mode, hooks, checker and approval requests. (b) No project file widens permissions before trust; a test walks every key the loader reads. [\cd{P1}]
\item $\star$ (a) Running a command without an approval request needs an exact reading. (b) Near-miss tables include control characters and dialect forms. [\cd{P2}]
\item For each denied path, renames, parents, links and sockets are tested from a sandboxed shell. [\cd{P3}]
\item (a) Every tool with a path parameter uses one hidden-path check, for reads too. (b) Every security change runs the ordinary-work set. (c) Whether a project may read its own secret files is a stated choice. [\cd{P4}]
\item (a) ``Watched'' is decided per turn. (b) Application text defaults to nobody. (c) Every ask path refuses at once when unwatched. (d) Commands that change settings are refused in unwatched turns. [\cd{P5}]
\item (a) Nothing a folder's files declare runs or grants before the person trusts it. (b) Extension grants count only in watched turns, and repository hooks are bound to a hash. (c) Another tool's allow rules never run a command unasked. [\cd{P6}]
\item (a) Approval requests show the harness's reason first. (b) They never answer themselves. (c) They are withdrawn at stop. (If requests carry ids, answers are matched by id: \Bel.) [\cd{P7}]
\item (a) Saved grants live where the model cannot write. (b) A sandbox-exit grant matches only the exact command. (c) Every refusal names its source. [\cd{P8}]
\item Outside text is fenced as data and never changes whether a turn is watched. [\cd{P9}]
\item[] \textbf{What the person sees}
\item $\star$ (a) Each event has one notice and one title. (b) Every number is a field from the record and counts what the person cares about. (c) Every final state has a reason. [\cd{V1}]
\item A helper's outcome, cost and stop reach the parent. [\cd{V2}]
\end{enumerate}
\end{multicols}
}

\section{A self-assessment of our harness}\label{sec:cross}

To show how the checklist is used, its 30 items were applied to Crewroom, the harness the cards come from, as one worked example. The assessment was done by the party that built the harness: the same author set the checklist and the scoring rule, and no second rater scored it. Two other harnesses were also run against the checklist during this work; their results are not reported, because they were mostly single runs under unequal conditions and their maintainers were not consulted. A few observations from those runs appear in the cards, marked as seen in other harnesses. The result here is therefore neither a comparison nor a ranking. Conditions: one-shot mode only, with display parts judged on machine-readable output; a scripted provider on one wire format, which for our harness is its generic route and not its main one; and most scenarios run once. Where an item about approvals, the command checker, protected paths, the sandbox, credentials or injected content does not hold, only the verdict is printed.

\paragraph{Set-up.} Our harness ran at one commit from a separate checkout, nothing installed. Each run had a new temporary home and configuration, a throwaway version-controlled work folder holding two files, an emptied environment, and an operating-system sandbox profile of the runner. Requests went through a local fault relay to a scripted provider of about 170 lines in the chat-completions format (text, tool calls, each finish reason, a refusal field, a reasoning field, a malformed event, a close with no finish reason, error statuses, a hang before headers); the relay adds overload, a rate limit with a retry-after, a cut, a stall and an empty reply. Silent-provider runs went to the scripted provider directly, because the relay gives up on a silent upstream after about 360\,s, and were capped at 1,000\,s. The harness ran in its one-shot mode with structured output and in its default mode, with nobody to ask; there it refuses every shell command that is not read-only, so scenarios that need a running command were run with a standing grant for the shell tool. Overload three times, the cut, one empty reply, the 60\,s rate limit, kill and stop ran three times, everything else once, including the timing results in rows F1 and F2. For kill, a tool ran a 30-second command, the harness was killed hard at 4\,s, and the session was resumed. For stop, a Ctrl-C was typed into a pseudo-terminal whose foreground process group held the harness: three runs mid-tool and one mid-stream.

\paragraph{Scoring rule.} Each lettered part of an item, as printed in Section~\ref{sec:checklist}, gets one verdict. \textbf{H}, holds: a run showed it, or, for a part about code or tests, a positive finding in source. \textbf{N}, does not hold: a run showed a counterexample, or a positive source finding shows the required thing absent. \textbf{D}, different form: the intent is met by a mechanism the wording does not name; a form the wording already allows counts as H. \textbf{U}, not tested: no run reached it, it was read in source only, or something was not found in source; not finding something is never N. \textbf{NA}: the card's heading excludes the harness, or the part concerns a feature it lacks. An item does not hold if any tagged clause of its card was shown false in a run of our harness, even where the checklist item's own wording is narrower. An item is NA when the whole card is; otherwise it takes its worst tested part, in the order N, D, H; untested parts do not lower it, and an item whose applicable parts are all U is U. All items are scored on one-shot output, display parts included; evidence from another mode is noted, not scored.

{\scriptsize
\renewcommand{\arraystretch}{1.03}
\begin{longtable}{@{}>{\raggedright\arraybackslash}p{0.05\linewidth}>{\raggedright\arraybackslash}p{0.93\linewidth}@{}}
\caption{The 30 checklist items in our harness, by the rule above. ``(source)'' marks a reading of code, not a run.}\label{tab:self}\\
\toprule
\textbf{Card} & \textbf{Verdict}\\
\midrule
\endfirsthead
\toprule
\textbf{Card} & \textbf{Verdict}\\
\midrule
\endhead
\bottomrule
\endfoot
T1 & \textbf{N}: a stream with no finish reason shown and stored as complete, exit 0; step limit named\\
T2 & \textbf{H}: two calls, the first failing: ids and order kept\\
T3 & \textbf{N}: a stream with no finish reason was shown and stored as complete, exit 0 (clause (b) Open); nothing ran from a cut reply and the cut text was not in the output (3 of 3)\\
T4 & \textbf{H}: reasoning not shown\\
T5 & \textbf{N}: a timed-out command ran on and finished; result ``timed out'' after 45\,s. Empty and large output held\\
F1 & \textbf{N}: 18 requests in six bursts of three, the request layer inside each of six rounds; the stop time shown is not the bound that ends it\\
F2 & \textbf{N}: one-shot output silent for 300\,s, then a reconnect row; not ended by 1,000\,s; stop time shown 10 or 60 minutes after the start, real end after 4 to 5 minutes (Section~\ref{sec:realruns})\\
F3 & \textbf{H}: a 60\,s retry-after waited 60.4\,s (3 of 3); no reset time invented; (c) not tested\\
F4 & \textbf{H}: re-asks shown with true counts; bounded at two waits (the give-up line has no count; it is not a listed event)\\
F5 & \textbf{N}: filtered call not run, notice, re-ask; but the refusal was reported as an empty reply and re-asked under an empty-reply notice, so the person is not told the model refused (clause (b) Open)\\
C1 & \textbf{N} (c): the prompt names two tools absent from the request's tool list (loadable on demand)\\
C2 & \textbf{H}: identical across requests and across kill and resume (3 of 3)\\
C3 & \textbf{U}: a context-length error led to a summary; the parts not checked\\
S1 & \textbf{H}: call stored before the kill; model told the outcome is unknown; the command died with the engine (3 of 3)\\
S2 & \textbf{N}: result ``aborted'' at once, but the command ran on and finished; exit 0 after about 30\,s (3 of 3); no new request\\
S3 & \textbf{U}: no input channel\\
S4 & \textbf{N} (source): 24 of the 65 parity scenarios do not run over the pipe\\
S5 & \textbf{U}: not run\\
S6 & \textbf{H}: every request that returned usage booked; unknown price not zero; a capped run on an unpriced model refuses to start\\
P1 & \textbf{N}: does not hold in our harness\\
P2 & \textbf{N}: does not hold in our harness\\
P3 & \textbf{U}: not tested\\
P4 & \textbf{N}: does not hold in our harness\\
P5 & \textbf{H}: every ask refused at once\\
P6 & \textbf{H}: the declared server not started before trust\\
P7 & \textbf{U}: no approval requests in the one-shot mode\\
P8 & \textbf{H}: a write to its own settings refused by a named safety check\\
P9 & \textbf{N}: does not hold in our harness\\
V1 & \textbf{N}: 18 requests counted at most ``2 of 3'' per round; a refusal announced as an empty reply; a command reported timed out although it finished\\
V2 & \textbf{H}: a helper's failure reached the parent with its reason, with its token count\\
\end{longtable}
}

\paragraph{Counts.} Recounted from the verdict column of Table~\ref{tab:cards} and from Table~\ref{tab:self}: of the 30 items, 11 hold (T2, T4, F3, F4, C2, S1, S6, P5, P6, P8, V2), 14 do not (T1, T3, T5, F1, F2, F5, C1, S2, S4, P1, P2, P4, P9, V1), none holds in a different form, 5 are not tested (C3, S3, S5, P3, P7) and none does not apply; 11 + 14 + 5 = 30. This is a self-assessment by the builder under the conditions above, and one counterexample makes an item N by the rule: F5 is N because the refusal was announced as an empty response, and T3, P4 and P9 are N for the same reason, each by a clause of its card that a run showed false. The 14 items that do not hold rest on twelve distinct observations, since one observation can fail several items (the stream with no finish reason fails T1 and T3; the command that ran on after its timeout, T5 and V1; the overload bursts, F1 and V1; the refusal announced as empty, F5 and V1). The harness is a private application that is not distributed, and the items that do not hold are being worked on.

\paragraph{Findings that weigh most.}
\begin{itemize}[leftmargin=1.2em,itemsep=1pt,topsep=2pt]
\item \emph{A stop that lets a command run on} (S2). After a Ctrl-C 4\,s into a 30-second command, the result said at once that the command was ``aborted before completion'', but the command ran on and finished its side effect, and the process waited for it and exited 0 after about 30\,s (3 of 3). No new request followed, and a stop mid-stream ended at once (1 of 1).
\item \emph{A timeout that does not end the command} (T5). A command with a 3\,s timeout was reported as ``timed out after 3s'', yet ran on, finished its side effect and held the turn for 45\,s (1 run).
\item \emph{A request count the person cannot see} (F1, V1). Under overload on every request our harness sent 18 requests in six bursts of three, separated by five waiting rounds: the request layer tries three times inside each of six rounds of the loop. The row counted ``1 of 3'' and ``2 of 3'' in the one-shot output (``1 of 10'' and ``2 of 10'' in the chat), starting again each round; the log said ``after 3 attempts''; the final line gave no number.
\item \emph{A stop time shown that is not the real one} (F2). The waiting row showed a stop time ten minutes after the start for about 230 of 244\,s, and the runs ended 5.0 to 5.2 minutes before it (3 of 3); the one-shot generic route named 60 minutes and ended after 286\,s (Section~\ref{sec:realruns}).
\item \emph{A prompt that names tools the request does not carry} (C1). The system prompt told the model to use two tools absent from the request's tool list; both exist as tools loaded on demand, and whether the model could have loaded them was not checked.
\end{itemize}
Its other items that did not hold are T1, T3, F5, S4, P1, P2, P4 and P9; for P1, P2, P4 and P9 only the verdict is given. P3 was not tested. It held in S6 and in S1: after a hard kill the command died with the engine and the model was told the outcome was unknown and to check it (3 of 3). In 15 of 88 one-shot runs processes started by tools were still alive one second after exit; how long they live was not measured, and the claims of T3(b) and S2 of no running process behind are not shown by these runs. The clauses these runs found false are Open in the cards: T3(b), T5(b), F1(b), F5(b), C1(a), F2(b), S2(d), S2(e), P1(a), P2, P4, P9(a) and V1(d).

\paragraph{What this check is and is not.} Section~\ref{sec:open} gives the general limits. The same relay and launcher sat in front of all runs, and the relay misled once: its upstream limit of about 360\,s, which the silent-provider runs then bypassed. Runs were made with an emptied environment, inside a sandbox profile of the runner, and without a terminal except for the stop runs; any of these can change a harness's behaviour. The check cannot show real outages, real rate limits or a model's response to injected text; the relay's fault shapes may be classed differently from real ones; interactive approval requests, mid-turn messages and the terminal displays were not seen; and deterministic scenarios ran once. The tags in the cards describe our harness alone. The raw evidence notes score four items (T3, F5, P4, P9) more favourably under a narrower reading of the checklist; this paper uses the stricter reading. For one item, F4, the paper's verdict is the more favourable of the two, because the printed wording does not list the give-up line.

\section{Limits and open questions}\label{sec:open}

The cards stop where the evidence stops. \emph{Limits of the defences.} The cards on the sandbox and on credentials state their limits as classes. Code running as the same user outside the sandbox can read anything the person can; only a separate machine closes that. 

\paragraph{Limits of the evidence.} \emph{One author.} The same author set the checklist and the scoring rule, and there was no second rater. No person outside the project reviewed the harness or re-ran its tests. \emph{Scripted providers.} The self-assessment used a scripted provider behind a relay, in one-shot mode. \emph{Few real runs.} Of 34 runs of our engine behind the relay, 28 reached one small real model of one family through one wire format (21 of 27 in the second session and all 7 in the first); in the other six the relay answered every request. No run used a second family. \emph{Real failures, counted.} A read-only script counted the engine's own event lines in the per-chat logs of one person's use over about three weeks (872 chat files, 4,569 event lines); nobody opened the files, and only fixed fields were read (Table~\ref{tab:failcount}). Retries were rare against completed turns, and a stop by the person was as common as a retry; the two counts cover different windows, because turn events were logged only from a later point than retries, so the comparison is rough. What was not countable: error and notice codes (give-up, cut-off, no answer, rate limited, offline) are written only since a late change and none appeared, so zero there means no coded line was written, not that it never happened; the 43 error rows carry no code, so their kind is unknown; wait rows go to the window and never to the log; a silent stream or a cut-off reply appears only through the labels in the table. The counts are events, not incidents, from one person's use while the application changed, so rates across versions are not comparable; deleted chats are absent, so all are lower bounds. \emph{Readings.} The ten harnesses were tested without versions or dates. \emph{No artefact.} The scripted provider, scenarios and runners are not published, and no table ties each Tested clause to a test and the date it last passed.

\begin{table}[tb]
\centering\footnotesize
\renewcommand{\arraystretch}{1.1}
\caption{Failure events in one person's real use of our harness, counted from the engine's own event lines. Events are not incidents; all counts are lower bounds.}\label{tab:failcount}
\begin{tabular}{@{}lrr@{}}
\toprule
\textbf{Event} & \textbf{Events} & \textbf{Chats}\\
\midrule
Turn completed & 320 & 160\\
Turn stopped by the person & 40 & 36\\
Request retried, all labels & 38 & 15\\
Switch to a backup model & 4 & 4\\
Model call ended aborted & 10 & 9\\
Model call ended in error & 13 & 5\\
Reply dropped & 2 & 1\\
Error row shown, no code field & 43 & 38\\
Stop at the step limit & 3 & 3\\
Stop by the loop guard & 1 & 1\\
Engine exit & 8 & 6\\
\bottomrule
\end{tabular}
\end{table}

\paragraph{Open questions.} Should silence switch to the backup? Which counters reset after an account switch? What deadline applies to chats in which the harness drives another agent program? Is there a better signal than wording lists? How many refusals should end a turn, and do models retry refused actions in another spelling? Should look-alike letters be normalised? Should the network be off by default? Should approval requests survive a restart and be answered by id? Should the final answer be checked against the tool outcomes? Do fences or prompt changes alter real behaviour? 

\paragraph{Not covered.} Answer quality, model choice, cost across models, visual design, phones, shared multi-person use and other operating systems are out of scope. Rewind, checkpoints, forks and saved memory have tested mechanisms but no card; paused replies, approval requests for parallel calls and per-tool timeouts have little evidence. 

\section{Related work}\label{sec:related}

\paragraph{Agent systems and scaffolds.} Many systems define an agent loop around a model and its tools~\cite{react,autogen,sweagent,openhands,agentless,magentic}; surveys and frameworks organise agents by component~\cite{wangsurvey,coala}; a study of practitioner talks maps the themes of building language-model applications for production~\cite{mailach}. Interface principles for agents have been tested by ablation~\cite{sweagent}.

\paragraph{Closest work and what this paper adds.} Four works are closest; we describe them as their abstracts do, without having read them in full. A pattern catalogue~\cite{liu} gives 18 architectural patterns for agents built on foundation models, with forces, trade-offs and a decision model; its unit is a pattern, not a decision with a test. A source-code taxonomy~\cite{rombaut} places 13 open-source coding-agent scaffolds, read at pinned versions, along 12 dimensions in three layers. A system paper~\cite{bui} presents the design of one open-source command-line coding agent. A deployment report~\cite{srinivasan} names three missing primitives of a tool protocol, sorts production failures into five design dimensions and gives a readiness checklist. This paper adds, for the whole harness, an invariant and a test for each decision, the corner cases that break it, a strength tag on every clause, and a worked self-assessment of the checklist on one harness.

\paragraph{Established techniques under other names.} Several terms here rename older ideas. Planted changes are mutation testing~\cite{demillo}; the behaviour lock is characterisation, or golden-master, testing~\cite{feathers}; the differential test of verdicts follows differential testing~\cite{mckeeman}; ``two readers'' is the parser-differential problem~\cite{sassaman}; ``borrowed authority'' is the confused deputy~\cite{hardy}; card S1 is write-ahead logging~\cite{aries} with the crash-consistency questions it raises for applications~\cite{pillai}; card F1's multiplying retries are the retry amplification behind metastable failures~\cite{bronson}; and the fake clock is a small case of deterministic simulation testing~\cite{fdb}.

\paragraph{Harness effects on results.} Several studies examine how the harness affects measured results~\cite{vats,hu,bensghaier}, one varying planning, action space and context management with the loop held fixed~\cite{fan}; this paper makes no performance claims.

\paragraph{Failures, security and context.} Coded failure modes~\cite{cemri}, failure attribution~\cite{zhang}, a production report with failure vignettes and a readiness checklist~\cite{srinivasan}, and red-teaming lessons~\cite{bullwinkel} are the models for Sections~\ref{sec:catalogue} and~\ref{sec:checklist}. Indirect injection~\cite{greshake}, its benchmarks~\cite{injecagent,agentdojo}, defences by construction~\cite{camel}, design patterns that trade utility against security~\cite{beurer}, emulated sandboxes~\cite{toolemu} and the agentic attack surface~\cite{sok} underlie Section~\ref{sec:perm}; context management~\cite{memgpt,lostmiddle,salle} relates to Section~\ref{sec:told}, and evaluation practice~\cite{swebench} to Section~\ref{sec:testing}. Design hints from building systems~\cite{lampson} and the end-to-end placement argument~\cite{saltzer} are the older model.

\section{Conclusion}\label{sec:conclusion}

A harness is mostly made of decisions about edges: how a turn ends, what happens to work in flight, what is written before it is shown, and who is allowed to decide. A few ideas recur across the cards. Every ending has a name, and everything that starts ends exactly once. Every wait and every retry has one limit that someone chose. Nothing untrue is said to the model or to the person, and a guess is never shown as a fact. When two components read the same thing (the checker and the shell, the prompt and the tool list, two transports), they must agree, or the stricter reading wins. Authority belongs to a turn and to the person's own files, not to a chat, a repository or another agent. And a test is evidence only if it can fail, is actually run, and runs the way real use runs. The evidence comes from building one harness, Crewroom, from manually testing ten others and studying open-source harnesses as reference, from 34 runs behind a fault relay, 28 of which reached one small real model, from a count of failure events in one person's use, and from a self-assessment of that harness against the checklist with a scripted provider and one scoring rule. Of its 30 items, 11 hold, 14 do not and 5 were not tested, on twelve distinct observations (one observation can fail several items); the harness is a private application that is not distributed, and the items that do not hold are being worked on; the assessment was made by the builder and is a worked example, not a measure and not a ranking. Several of our invariants are design choices and are written as such; and in our harness a Stop and a timeout reported commands as ended that ran on, and the waiting row named a stop time five to six minutes after the real end. The cards are offered as a checklist to test against, not as proof that any harness is correct.


\appendix
\section{Evidence notes per card}\label{app:evidence}

Each note traces the evidence behind a card's tags: what was observed, and on what kind of model or set-up. ``Scripted'' means a scripted provider; ``on screen'' means checked in the installed application; ``real'' means a real model provider. Notes say whether a point was seen once or observed in other harnesses. Counts and timings in the cards and here that are not in the recorded runs or the tables (the ten-run counts of F4, the baseline and file counts of S4, the timing and size of P3, the caller count of P5, the case count of P6, the string counts of the P cards, the thresholds and the step count observed in other harnesses or noted during development) are from the author's notes.

\begin{small}\begin{description}[leftmargin=1.2em,itemsep=1.5pt,topsep=2pt,parsep=0pt,font=\normalfont\bfseries]
\item[T1.] A script over our harness's turn function lists 62 exits. At the last count the behaviour lock reached all 62 (61 through its fault rows); the parity recording alone reached 16 (Section~\ref{sec:repeat}). A stream with no finish reason was shown and stored as complete in the one-shot runs of Section~\ref{sec:cross}. The closing-message rules were scripted. Cut-reply miscounts were seen once, in a real run. 
\item[T2.] One test runs nine scripted scenarios and asserts one start and one end per call; the code has 21 outcomes in 13 classes (Table~\ref{tab:endings}), not each with its own scenario. 
\item[T3.] Early reads were scripted with a stalled reply; drops with a read in flight were injected; all three output modes were checked. Keeping early results for a retry was observed in no harness. 
\item[T4.] The opening sentence stored after the call, and drawn as reasoning, was seen during development on two model families on one wire format, from the author's notes and not recorded; arrival order and the reasoning shapes are scripted. The step-group layout is from comparison. The one rendering test does not prove that a repeated text is drawn only once. 
\item[T5.] Size cap, saved output, empty-output sentence, unchanged re-reads, backgrounding past a timeout and result clearing were scripted; the once-a-second stop polling is thinly tested, and redaction of saved secrets has no test. Ending a watched turn for background work is from comparison. 
\item[F1.] A table test of the failure function and a fake-clock test bounding the total wait passed; a dozen situations were still decided differently in two places. Real runs: Section~\ref{sec:realruns}; request counts in the one-shot runs: Section~\ref{sec:cross}. 
\item[F2.] The deadline, waiting line and final line were checked with a fake clock and a failing scripted provider in the installed application; the waiting line was seen in one chat driven through an outside agent program and in real runs; under overload on every request, in runs the relay answered alone, the stop time in the waiting row was five to six minutes after the real end, so clause (b) is false (Section~\ref{sec:realruns}); no real outage was seen. Ending without waiting rounds once text was shown was tested with a fault injected after text. A false ``answered after N tries'' row was seen once in a real run. 
\item[F3.] Classification by recorded error bodies, stated-only reset times, account order, backup request shape, the switch row, the label and the return were scripted on each wire format; the switch row and label were seen live. In real runs a rate limit with a 60\,s retry-after was waited out and retried, worded as overload; a longer retry-after was not tried. Wording lists are the weak point. 
\item[F4.] Empty-reply, announcement, repeat, repetition, repair and request-field guards were scripted, because they cannot be forced on real models; empty-reply rows were checked on screen with a scripted model; no repair fired in real runs. Ten real runs per option on one small model: duplicate closings in 3 of 10 before, 5 of 10 with the prompt sentence, 4 of 10 with prompt and reworded tool result, 0 of 10 with the display rule (seen once). Behind the relay an empty reply on one family was not re-asked and the turn failed with a plain line; this is a coded rule for that family, confirmed on three empty shapes with a scripted upstream (9 of 9), with one nudge when the empty reply follows tool results (3 of 3; Section~\ref{sec:realruns}). 
\item[F5.] Refusal and filter paths were scripted with and without calls; no real refusal was forced. 
\item[C1.] Prompt-assembly unit tests per family and chat kind pinned the removal of untrue and contradictory claims and the tool-name filter; the helper-role and family-selection tests are planned, not run; live behavioural checks have no results yet. 
\item[C2.] Prefix-stability tests on recorded request bodies passed for one vendor's direct route; the restart test and the size ceiling were planned; the cost of a volatile value was computed, its frequency not measured; gateway routes send no markers. Posture switching is from comparison. 
\item[C3.] Typed-message retention, rule replay, the measured-facts check with one repair, and the bounded overflow resend were tested; clicked decisions, attachments and the empty list after restart were found by probing chat shapes (seen once). 
\item[S1.] Kill-and-resume tests at fixed points and a two-engine lock test passed. Read from code: writes are appends not forced to the device; a torn last line is ended with a newline at the next start of a served engine and skipped on read (the notes name a test for it, not re-run here); a call that never started before a crash reads ``outcome unknown'' (clause (d)). Cutting the log back and automatic continuation are from comparison. 
\item[S2.] Stop at each phase (thirty repeats for the open-approval and slow-hook phases), result texts per cause, withdrawal of approval requests, helper stop text, exit codes and hooks were scripted. One stop test failed once in ten grouped runs with no cause found. In 15 of 88 one-shot runs of our harness processes started by tools were still alive one second after exit (Section~\ref{sec:cross}). 
\item[S3.] Steering at each phase and resend after a restart were scripted; a just-sent message shown twice was seen once on screen. 
\item[S4.] The behaviour lock runs the 41 pipe scenarios against their own baselines and was green (Section~\ref{sec:repeat}); that checks each transport against itself, not the two against each other. 
\item[S5.] Flow control was checked by a probe; the crash-loop limit and the handshake were scripted. The headless-server target and numbered events for reconnect are from comparison. 
\item[S6.] Booking once per billed request, unpriced models, the budget stop and the usage correction were scripted. Hidden checker and memory calls missing from task cost were seen once. 
\item[P1.] Precedence matrices, loader-key planting, protected-path writes and the credential matrix were scripted. The harness-held mode for outside agents was found in a comparison study and not built. 
\item[P2.] A review pass found a whitespace-class misreading after a loosening (case~1); it was closed by the control-character rule; near-miss tables of hundreds of strings per approved form and a dialect corpus of over a thousand strings run in every mode. The checklist item for this card does not hold in our harness (Section~\ref{sec:cross}). 
\item[P3.] Denies for read, rename, parents, links and sockets were tested from a sandboxed shell; a link-following step opened a hole and was removed; start-up cost per command was measured. The rule for paths that do not exist yet is from comparison. Each deny was tested on its own; the profile as a whole is not confirmed by a second check. The checklist item for this card was not tested in the self-assessment runs. 
\item[P4.] A registry sweep over every path-taking tool, name-variant lists and a fake-home tool check passed; the ordinary-work breakages were found only when ordinary work was probed (case~2); secret-named variables are removed from command environments, checked with canary variables. Each rule was tested on its own; the rules together are not confirmed by a second check. The checklist item for this card does not hold in our harness (Section~\ref{sec:cross}). 
\item[P5.] Per-turn watching, application-text defaults (a table of 23 callers), settings-changing commands and child runs were tested; one ask path (a sign-in request) has no test. Each rule was tested on its own; the rules together are not confirmed by a second check. 
\item[P6.] Skill-grant folder checks, hook trust by hash, linked-checkout inheritance (a twelve-case table), server pinning and the import rule were scripted. The imported environment block was found by a probe in a review pass (seen once). Hook-script hashing and model-judged hooks are untested. 
\item[P7.] The separate reason row, escaped characters, one-tap limits, withdrawal at stop and reminders were tested; answering by id is a written design; a request's state across a restart was tested, and the text the model then received was found wrong (the test is recorded as failing). 
\item[P8.] Attempts to write the grant file from file tools and a sandboxed shell in every mode, byte-variant matrices for sandbox-exit grants, and refusal texts per path were tested; the breaker exists, and its count was chosen without evidence. 
\item[P9.] Stripping, repeat refusal, hop limits and fence presence were scripted (three fence tests); behaviour under injection was not measured. The checklist item for this card does not hold in our harness (Section~\ref{sec:cross}). 
\item[V1.] A notice helper with a source check, row counts and title tables were scripted; the row verb was tested for a stop, not yet for a denial; that the error notice and the chat's status area show the same title was not checked on screen; several rows were found only on screen. 
\item[V2.] Helper status, stop text and cost booking were scripted; the wrapped-tool defect was observed in another harness, not reproduced. 
\end{description}\end{small}

\section{Two tables the cards point to}\label{app:tables}

General forms of two tables in our harness's code, without names, files or exact strings; numbers are our harness's choices. The code's 21 distinct call endings fall into the 13 classes of Table~\ref{tab:endings}.

{\footnotesize
\renewcommand{\arraystretch}{1.0}
\begin{longtable}{@{}>{\raggedright\arraybackslash}p{0.03\linewidth}>{\raggedright\arraybackslash}p{0.13\linewidth}>{\raggedright\arraybackslash}p{0.22\linewidth}>{\raggedright\arraybackslash}p{0.32\linewidth}>{\raggedright\arraybackslash}p{0.2\linewidth}@{}}
\caption{How a tool call can end (card T2): every call the person saw start ends in exactly one of these, matched by call id.}\label{tab:endings}\\
\toprule
\textbf{\#} & \textbf{Ending} & \textbf{Cause} & \textbf{What the model is told} & \textbf{What the person sees}\\
\midrule
\endfirsthead
\toprule
\textbf{\#} & \textbf{Ending} & \textbf{Cause} & \textbf{What the model is told} & \textbf{What the person sees}\\
\midrule
\endhead
\bottomrule
\endfoot
1 & Finished & The tool ran & Its output, never empty (say ``no output''), bounded; the full text kept retrievable & Done, with the result\\
2 & Failed & The tool ran and raised or reported an error & The error text & Failed, with the text\\
3 & Malformed call & Arguments unparseable, unknown tool, or tool not yet loaded & That nothing ran, why, and how to send it again & Failed row\\
4 & Refused by a limit & A run-level rail (calls, time, spend) & That it did not run, and to wrap up with what it has & Failed row naming the limit\\
5 & Refused by an extension & A hook or plug-in vetoed it & The veto text; a veto may also halt the whole run & Failed row; a stop notice if halted\\
6 & Refused by mode & The current mode does not allow it & That permission was denied, other tools may be tried, and the person decides & Failed row\\
7 & Refused by a guard & A built-in guard (repeat check, protected own record, special gates) & What is protected and what to do instead & Failed row\\
8 & Refused by permission & A rule, a session with no prompts, a safety check, or the person said no & Who refused, and that it did not run; ``nobody declined'' when nobody did & Failed row; the request card shows its outcome\\
9 & Request closed unanswered & Stop, closing the chat, or a helper stopped while the approval request was open & That it did not run and nobody refused it, so it may be asked again & ``Didn't run X'', with the reason\\
10 & Interrupted & Stopped before the call started, or the tool was aborted & One fixed interrupted sentence & ``Didn't run X'', stopped\\
11 & Discarded & The reply that made it is being retried & That it was discarded because the reply is retried & Row removed, no trace\\
12 & Not run, reply lost & The reply was cut, failed, stopped by a safety check, or hit the spending cap before the call started & A distinct sentence per cause & ``Didn't run X'' with a distinct status\\
13 & Unknown after a crash & The process died before the result was written & That the outcome is unknown and must be checked before repeating & Failed row on resume\\
\end{longtable}

\begin{longtable}{@{}>{\raggedright\arraybackslash}p{0.03\linewidth}>{\raggedright\arraybackslash}p{0.15\linewidth}>{\raggedright\arraybackslash}p{0.19\linewidth}>{\raggedright\arraybackslash}p{0.32\linewidth}>{\raggedright\arraybackslash}p{0.21\linewidth}@{}}
\caption{From failure to action (card F1), 23 rows. The deciding function returns one of the eight verdicts listed in F1; request-layer retries, repairs, the stall rule and the refusal and filter re-asks are decided outside it.}\label{tab:errors}\\
\toprule
\textbf{\#} & \textbf{Failure} & \textbf{How recognised} & \textbf{What the harness does} & \textbf{What the person reads}\\
\midrule
\endfirsthead
\toprule
\textbf{\#} & \textbf{Failure} & \textbf{How recognised} & \textbf{What the harness does} & \textbf{What the person reads}\\
\midrule
\endhead
\bottomrule
\endfoot
1 & Bad credentials & Status 401 or 403 & Never retry; refresh a login token once, re-key once; then stop & Which credential is bad and where to fix it\\
2 & Model not allowed & 403 or 400 naming the plan or model & Never retry & The model is not on the plan; pick another\\
3 & Bad request & 400 & Never resend the same bytes; one repair per known cause; resend once if the error names a limit & The provider's words\\
4 & Unprocessable payload & 413 or 422; image size & Shrink or strip once; then stop & What was too large\\
5 & Context too long & The error names the context limit & Summarise and resend, at most three times & A final ``too long, start fresh'' line\\
6 & Output cap too large & The error names a limit & Resend once with the stated limit & Nothing unless it recurs\\
7 & Spent credit & 402 or billing words & Never retry & Add credit or pick another model\\
8 & Spent plan & 429 with a spent-plan body, a reset over an hour, or daily or weekly wording & Never retry; next account, then backup, else stop & The reset time, only if the provider stated it\\
9 & Short rate limit & 429 with a wait of a minute or less & Wait the stated time; retry within a small budget & ``Busy, trying again in N\,s (n of m)''\\
10 & Long rate limit & A wait over a minute & Do not wait it out silently; switch or stop & ``Try again in about X, or pick another model''\\
11 & Overload & An overload status or ``overloaded'' & Retry with backoff; after two or three tries switch to a backup; else up to five waiting rounds & A busy row, then ``still busy, stops trying by T''\\
12 & Server error & Other 5xx & As overload; with a backup, switch at once & An error row with a countdown\\
13 & Offline & Connection failure to a public host & Retry until the deadline, backoff growing to a minute & ``No internet, waiting; stops trying by T''; the final line says when it stopped\\
14 & Refused by a private host & Connection failure to a local address & Treat as a drop; small budget & ``Cannot be reached''\\
15 & No first byte & No headers within a time that grows with request size & One to five retries by family, with a longer window after & ``Did not answer for X, trying once more''\\
16 & Stall mid-stream & Silence for a stated time, heartbeats counting as life & Keep what arrived, or reconnect once or a few times & ``Went quiet for X, reconnecting''\\
17 & Dropped mid-reply & Reset or early close & Retry with backoff within a budget; one extra send past the deadline if text was shown & ``Connection dropped, trying again (n of m)''; ``cut off n times''\\
18 & Cut inside a tool call & Length stop with a half call & Resend with more room, at most three times & A row naming the resend\\
19 & Empty reply & No text and no calls & By family: resend after 5\,s then 10\,s, or nudge once; for one family an empty first reply that does not follow tool results ends the turn at once (Section~\ref{sec:realruns}) & ``Empty answer, asking again'', or a plain line naming the next step\\
20 & Safety refusal & Refusal stop reason & Ask once with a note; do not run its calls & ``Safeguards stopped the response; continuing once with that noted''\\
21 & Content filter & Filter stop reason & Ask up to five times with backoff & ``Content filter blocked the answer; continuing without that part (n/5)''\\
22 & Deadline & Time since the first failure passes ten minutes (watched) or an hour (unwatched) & Stop; one extra send if text was shown & ``Stopped trying at T; your message and work are saved; send continue''\\
23 & Switch & After rows 8 and 10 to 12 & Account, then backup; reset counters by named lists; label every backup reply; return only when the account is free & A switch row and a label (F3)\\
\end{longtable}

\paragraph{Approval rules (card P2).} Our harness reads each shell command twice, with a strict reading used to approve and a permissive reading used to refuse and to match deny rules, and it asks the person whenever the strict reading cannot model a command exactly; the grammar is not printed.

}

\section{Glossary}\label{app:glossary}

{\scriptsize
\begin{multicols}{2}\raggedright
\begin{description}[leftmargin=1.2em,itemsep=0pt,topsep=0pt,parsep=0pt,font=\normalfont\bfseries]
\item[Allow, ask and deny rules.] Rules that let an action run, require an approval request, or refuse it.
\item[Application text.] Text the harness itself sends to start or continue a turn (a nudge, a restart prompt).
\item[Backup hold.] The time a chat stays on a backup model after a limit, until the main model's reset.
\item[Backup model.] A model the harness switches to when the chosen one is out of quota or unavailable.
\item[Endpoint.] One entry point of the harness's own interface, such as a settings route, through which a program can act.
\item[Fence.] A marker around outside content saying it is data, not instructions.
\item[Finishing tool.] A tool the model calls to signal it is done, after which it writes its closing.
\item[Folder trust.] The person's statement that a project folder's own settings, hooks and servers may take effect.
\item[Gateway.] A service that forwards requests to several providers through one wire format.
\item[Heartbeat.] A small message a provider sends during a long silence to show the connection is alive.
\item[Hook.] A program the person configures to run at fixed points, such as before a tool or at session end.
\item[Key agent.] A background program that holds login keys and signs with them through a socket, without exposing a file.
\item[Linked checkout.] A second working copy of the same repository that shares its history.
\item[Mode.] The person's chosen level of what may run without asking.
\item[One-shot run.] A non-interactive run that takes one prompt, prints a result and exits.
\item[Outside agent program.] Another agent program that the harness starts and drives inside its own window.
\item[Peer.] Another agent that sends messages into a chat.
\item[Probe.] A saved input with its expected verdict, used to show a hole and later that it is closed.
\item[Protected path.] A file that defines rules or the harness's own state (settings, hooks, tool-server configuration), which no tool may write without the person.
\item[Provider.] A service that runs a model and answers requests; a \emph{route} is one path to it (direct or through a gateway).
\item[Registry.] The list of every registered tool, command or settings key, which tests can walk entry by entry.
\item[Retry-after.] A provider's statement of how long to wait before trying again.
\item[Row.] One entry in the chat display. A \emph{notice} is a row that announces an event; a \emph{line} is a row of plain text from the harness, such as the waiting line.
\item[Session.] A chat together with the settings it runs under; what is true for a session depends on both.
\item[Skill.] A file of instructions, and possibly allowed tools, that the model can load for one kind of task.
\item[Spent-account record.] A shared note that an account is out of quota until a stated time.
\item[Step limit.] The most steps one turn may take before it stops and says how to continue.
\item[Stop reason.] The provider's statement of why a reply ended (finished, wants a tool, cut, paused).
\item[Surface.] A place where the person can answer, such as the main window, a phone or a notification.
\item[Tool server.] A separate program that offers extra tools to the harness through a standard protocol.
\item[Waiting round.] One long wait after a run of quick retries has failed, after which the retries start again.
\item[Weak-model guard.] A bounded runtime check that re-asks, nudges, stops or repairs a bad reply shape.
\end{description}
\end{multicols}
}

\end{document}